\documentclass{article}

\PassOptionsToPackage{numbers, compress}{natbib}
\usepackage{graphicx}
\usepackage{enumitem}
\usepackage{graphicx}
\usepackage{multirow}
\usepackage{booktabs}
\usepackage{subcaption}
 \usepackage[preprint]{neurips_2026}

\usepackage{wrapfig}
\usepackage[utf8]{inputenc} % allow utf-8 input
\usepackage[T1]{fontenc}    % use 8-bit T1 fonts
\usepackage{hyperref}       % hyperlinks
\hypersetup{hidelinks}
\usepackage{url}            % simple URL typesetting
\usepackage{booktabs}       % professional-quality tables
\usepackage{amsfonts}       % blackboard math symbols
\usepackage{nicefrac}       % compact symbols for 1/2, etc.
\usepackage{microtype}      % microtypography
\usepackage{multirow} 
\usepackage[table]{xcolor}
\usepackage{amsmath}

\title{Exposing and Mitigating Temporal Attack in Deepfake Video Detection}

\author{%
  \textbf{Zheyuan Gu\textsuperscript{1}\thanks{Equal contribution.}\hspace{1.15em}%
  Minghao Shao\textsuperscript{2}\footnotemark[1]\hspace{1.15em}%
  Zhen Wang\textsuperscript{3}\hspace{1.15em}%
  Keyu Mao\textsuperscript{4}\hspace{1.15em}%
  Ailiang Lin\textsuperscript{4}}\\[-1pt]
  \textbf{Mingyou Liang\textsuperscript{5}\hspace{1.15em}%
  Mingkun Xu\textsuperscript{6}\hspace{1.15em}%
  Shijie Zhang\textsuperscript{1}\hspace{1.15em}%
  Hao Jiang\textsuperscript{1}\hspace{1.15em}%
  Yusong Wang\textsuperscript{7}\thanks{Corresponding author.}}\\[2pt]
  {\normalfont\small \textsuperscript{1}Peking University\hspace{0.8em}%
  \textsuperscript{2}New York University\hspace{0.8em}%
  \textsuperscript{3}Huzhou Normal University}\\[2pt]
  {\normalfont\small \textsuperscript{4}Institute of Science Tokyo\hspace{0.8em}%
  \textsuperscript{5}Tongji University}\\[2pt]
  {\normalfont\small \textsuperscript{6}Guangdong Institute of Intelligence Science and Technology}\\[2pt]
  {\normalfont\small \textsuperscript{7}Hangzhou Tuomai Information Technology}
}

\begin{document}

\maketitle

\begin{abstract}
While spatiotemporal deepfake detectors achieve high AUC, our experiments reveal their susceptibility to evasion attacks. 
These models tend to overfit on fragile temporal spectrum cues, rather than learning robust semantic causality. 
To mitigate this vulnerability, we propose SpInShield, a temporal spectral-invariant defense framework explicitly designed to decouple semantic motion from manipulatable spectral artifacts.
We propose a learnable spectral adversary that dynamically synthesizes severe spectral deformations, simulating extreme attack scenarios. 
By employing a shortcut suppression optimization strategy, SpInShield compels the encoder to extract reliable forensic cues while purging unstable spectral statistics from the latent space. 
Experiments show that SpInShield obtains competitive performance on widely used datasets and outperforms the strongest baseline by 21.30 percentage points in AUC under simulated amplitude spectral attacks.
\end{abstract}

\section{Introduction}
\label{sec:intro}

The rise of hyper-realistic AI-generated faces (deepfakes) undermines digital media credibility \cite{Mirsky2020TheCA}.
This technology lowers the barrier for malicious actors to fabricate convincing videos and biometric bypass and political disinformation campaigns \cite{Chesney2018DeepFA, Tolosana2020DeepFakesAB}. 
These capabilities can destabilize digital media's evidentiary value, demanding robust, automated countermeasures.
In response, the community has deployed various detection mechanisms, mainly leveraging deep learning models to distinguish synthetic videos from authentic videos \cite{Verdoliva2020MediaFA}. 
Because dynamic videos contain temporal cues distinct from static images, modern defenses adopt data-driven architectures that mine spatiotemporal representations, making them the primary defense against digital manipulation \cite{Yan2024OrthogonalSD,pixelwise,FakeSTormer}.

Existing deepfake detection methods fall into static frame-level analysis and dynamic video-level modeling \cite{Verdoliva2020MediaFA,Masood2021DeepfakesGA}.
The first category extracts individual frame artifacts, treating video inputs as independent images. 
They use CNN backbones or spectral transformations to capture pixel-level anomalies and spatial frequency fingerprints, relying solely on intra-frame features \cite{FF++,Liu2021SpatialPhaseSL,10.5555/3737916.3739345}.
The second category exploits temporal correlations via spatiotemporal coherence, overlooked by static analyzers. 
3D-CNNs \cite{Tran2014LearningSF} or Video Transformers \cite{Arnab2021ViViTAV} analyze motion continuity and inter-frame relationships, revealing spatiotemporal inconsistencies that expose manipulation \cite{FTCN,Yan2024GeneralizingDV,FakeSTormer,pixelwise}.

\begin{wrapfigure}{r}{0.36\textwidth}
    \centering
    \vspace{-0.5em}
    \includegraphics[width=0.35\textwidth]{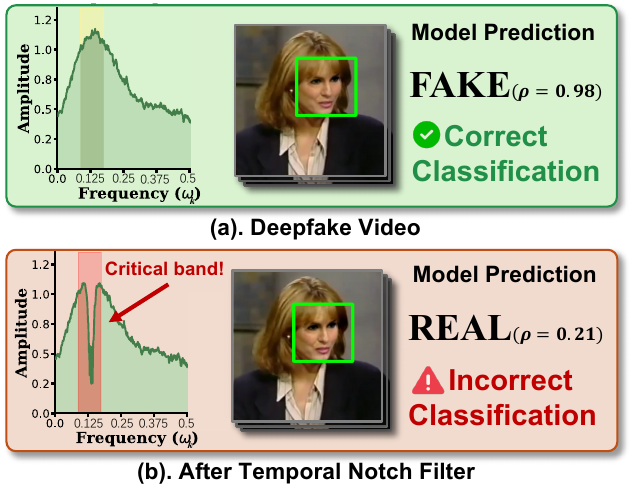}
    \caption{SLF~\cite{stylelatentflow} relies on temporal spectral cues for detection, and fails with misclassifications when these cues are suppressed.}
    \label{fig:firstimage}
    \vspace{-2mm}
\end{wrapfigure}

Despite high performance on standard benchmarks, these spatiotemporal models risk overfitting specific temporal spectral patterns.
\autoref{fig:firstimage} shows that selectively suppressing specific temporal frequency bands with a temporal notch filter can substantially reduce detection performance, measured by the area under the receiver operating characteristic curve (AUC), indicating sensitivity to these spectral components.
Whether implicitly modeled by neural networks or explicitly extracted via Discrete Fourier Transform (DFT), this rigid dependence on fixed temporal patterns makes the detectors vulnerable. 
The influence on the temporal spectrum, such as the aforementioned suppression, spectral tilt, and codec distortion, is common in the real world, arising from temporal filtering, global denoising, and video re-encoding.
In such scenarios, an adversary can easily employ various perturbations to modify these temporal spectrum patterns and evade detection, raising significant security concerns.

However, building robust detectors faces three challenges. 
First, separating malicious temporal spectrum artifacts from legitimate motion is complex. 
The temporal frequency of forgeries overlaps with genuine facial dynamics, such as micro-expressions or blinking. 
Naively suppressing specific frequency bands to avoid overfitting risks losing critical semantic information, reducing detection AUC on real videos.
Second, modern architectures exhibit a strong inductive bias toward low-level statistics. 
Models like 3D-CNNs and Transformers are driven by optimization to find the easiest solution.
Without explicit constraints, they naturally gravitate towards brittle temporal spectrum shortcuts rather than learning the complex, high-level semantic causality of video content.
Third, capturing spectrum-invariant representations requires a shift in learning paradigm. 
The lack of ground-truth annotations on spectral robustness prevents direct supervision to penalize the model for relying on brittle features. 

To counter temporal spectral attacks and address these challenges, we propose \textbf{SpInShield} (Temporal Spectral Invariance Protection), a spectral invariance defense framework.
First, to separate malicious spectrum artifacts from legitimate motion, we introduce a Learnable Spectral Adversary (LSA).
This generative mechanism synthesizes severe spectral amplitude deformations while preserving the phase of nonzero Fourier coefficients before clipping, simulating worst case attacks.
Second, to overcome the inductive bias toward low level statistics, we employ a weight sharing Siamese architecture.
By processing the original videos and their spectrally attacked counterparts simultaneously, it forces the encoder to discard brittle spectrum shortcuts and extract robust forensic cues within a shared representation space.
Finally, to achieve spectrum invariant representations without labels, we introduce a shortcut suppression optimization strategy.
By enforcing adversarial spectral blindness at the representation level and symmetric prediction invariance at the prediction level, this optimization autonomously purges environment specific spectral statistics from the latent space.
As a result, the model mitigates frequency shortcuts and maintains consistent classification decisions under temporal spectral attacks.
In summary, contributions of this work are as follows:
\begin{itemize}[leftmargin=*,noitemsep,topsep=0pt]

\item We identify that existing spatiotemporal deepfake detectors rely on amplitude-specific shortcuts in the temporal spectrum, creating an exploitable attack surface to evade detection.

\item We propose SpInShield as a spectral-invariant detector by learnable adversarial amplitude spectrum synthesis and shortcut suppression optimization, reducing reliance on unstable spectral statistics.

\item Experiments under simulated temporal spectral attacks demonstrate that SpInShield outperforms existing methods with a 21.30 percentage-point average AUC gain, validating its effectiveness.

\end{itemize} 

\vspace{-0.2cm}

\section{Background}
Existing deepfake detection methods operate primarily at the frame or video level. Methods at the frame level focus on spatial artifacts and high frequency anomalies \cite{FF++, F3Net, ProDet} but fail to model temporal dynamics explicitly. 
Detectors at the video level address this by capturing spatiotemporal inconsistencies \cite{altfreezing, FakeSTormer, stylelatentflow}, frequently relying on temporal frequency statistics. 
However, temporal frequency is susceptible to corruptions and adversarial perturbations \cite{benchmarking, cv_freq}. 
Driven by shortcut learning \cite{Geirhos2020ShortcutLI}, neural networks tend to exploit these spectral artifacts to minimize training loss rather than learning semantic causality \cite{LeveragingFrequencyForDeepfake}. 
Consequently, spatiotemporal encoders often regress to unstable spectral dependencies, resulting in generalization degradation under distribution shifts \cite{8639163}.  

\begin{figure*}[h]
    \centering
    \includegraphics[width=1\linewidth]{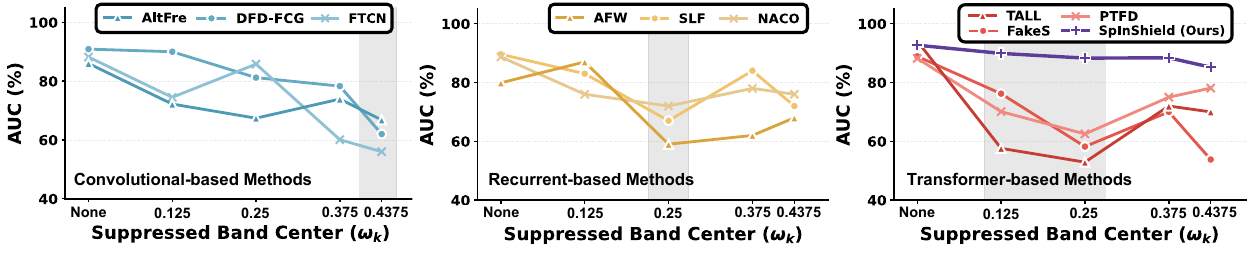}
    \caption{
    AUC under temporal notch suppression at representative DFT-bin frequencies.
    The x-axis shows the no-suppression and selected suppressed band centers $\omega_k=k/T$.
    Gray regions indicate the most vulnerable bands, where existing detectors degrade severely, while SpInShield remains stable.
    }
    \label{fig:teaser}
    \vspace{-4mm}
\end{figure*}

\vspace{-0.2cm}

\section{Preliminary}
In video analysis, temporal frequency analysis decomposes facial dynamics into two distinct components: \textbf{amplitude} and \textbf{phase} spectra.
Amplitude represents the energy distribution of a temporal signal and reflects low-level statistical attributes such as global illumination shifts and compression artifacts \cite{phaseandpowerspectra,1456290,dzanic2020fourierspectrumdiscrepanciesdeep}.
In contrast, phase encodes the temporal ordering and continuous semantic dynamics in facial motion, such as lip synchronization and blinking \cite{hommos2018usingphaseinsteadoptical}.

Given a clip of $T$ frames sampled at $fps$, where $fps$ is set to 25 by default, we partition the facial ROI into $M$ patches and compute the mean intensity of each patch to obtain temporal signals $x_m(t)\in\mathbb{R}$.
The patch-wise mean intensity is not used as a detection feature; it only parameterizes a phase-preserving spectral transform that generates the environmental view.
We apply DFT to obtain:
\begin{equation}
X_m(\omega_k)=A_m(\omega_k)e^{jP_m(\omega_k)}, 
\quad \omega_k=k/T,\ k=0,\dots,\left\lfloor\frac{T}{2}\right\rfloor,
\end{equation}
where $A_m(\omega_k)$ and $P_m(\omega_k)$ denote amplitude and phase, respectively.
Both phase and amplitude can be discriminative for deepfake detection.
However, artifacts from forgery generators typically manifest in the amplitude spectrum as statistical anomalies and other low-level spectral features \cite{Rubinstein2014AnalysisAV,10.1145/2185520.2185561,F3Net}. 
Compared with learning about the complex semantics like facial motion dynamics and micro-expression causality, models can more easily learn amplitude statistics directly \cite{10.1145/2461912.2461966,NGUYEN2022103525}. 
Moreover, for real-world attackers, manipulating phase while preserving realism is costly \cite{WANG2024104072,10.1145/1161366.1161375}.
We therefore hypothesize that current detectors rely primarily on amplitude-based shortcuts, making the amplitude spectrum the main source of overfitting and the most vulnerable attack surface.

\section{Empirical Analysis}
In this section, we address a central question: Do existing temporal-based detectors exhibit shortcut learning on amplitude-specific features?
We use the widely adopted FaceForensics++ (FF++) dataset \cite{FF++} (c23) and report the AUC.
To cover diverse temporal modeling paradigms, we evaluate nine representative methods, grouped by architecture:
\textit{Convolution-based Methods}: use 3D conv./temporal shift to extract spatiotemporal features, including DFD-FCG~\cite{dfdfcg}, FTCN~\cite{FTCN}, AltFreezing (AltFre)~\cite{altfreezing};
\textit{Recurrent-based Methods}: rely on Recurrent Neural Networks to model frame-to-frame continuity, including SLF~\cite{stylelatentflow}, NACO~\cite{NACO}, AFW~\cite{AFW};
\textit{Transformer-based Methods:} use self-attention to capture global temporal dependencies, including FakeSTormer (FakeS)~\cite{FakeSTormer}, PTFD~\cite{pixelwise}, TALL~\cite{TALL}.
% Per Eq.~(1), notch mask $M_{\text{notch}}(\omega_k)$ is applied to amplitude spectrum with phase fixed:
To quantify the sensitivity illustrated in ~\autoref{fig:firstimage}, we evaluate temporal notch suppression across detector architectures, with results reported in ~\autoref{fig:teaser}. 
Using the spectral representation in Eq.~(1), we apply a notch mask $M_{\mathrm{notch}}(\omega_k)$ to the amplitude spectrum while retaining the original phase terms:
\begin{equation}
\tilde{x}_{m}(t)
    = \mathcal{F}_t^{-1}\!\left( \big(A_m(\omega_k) \odot M_{\text{notch}}(\omega_k)\big)\, e^{j\, P_m(\omega_k)} \right),
\end{equation}
where $M_{\text{notch}}(\omega_k)$ specifies stopbands on discrete DFT bins.
We evaluate representative suppressed DFT-bin frequencies $\omega_k=k/T$.
Experimental results reveal a shortcut learning tendency: their temporal modules rely on low-level features of amplitude spectrum. 
Specifically, Recurrent-based methods degrade the most, with AUC dropping by up to 41.2\% relative to the no-suppression baseline.
This likely occurs as their temporal modules tend to track artifacts in specific frequency bands.
Similarly, Convolution-based methods and Transformer-based methods also drop by up to 36.8\% and 30.6\%, respectively. 
Thus, models may fail when the amplitude spectrum is attacked.

\begin{figure*}[t]
    \centering
    \includegraphics[width=0.95\linewidth]{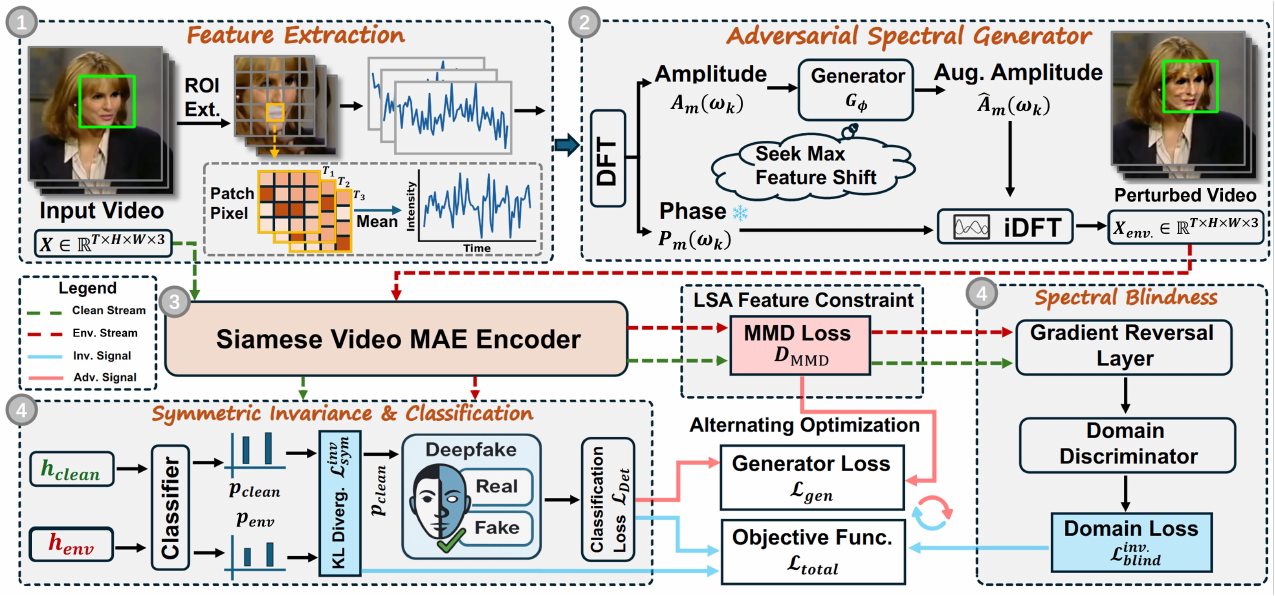}
    \caption{The framework comprising four interconnected modules: Feature Extraction, Adversarial Spectral Generator, Siamese Video MAE Encoder, and Shortcut Suppression Optimization.}
    \label{fig:structure}
    \vspace{-0.1cm}
\end{figure*}

\section{Methodology}

This section presents SpInShield shown in \autoref{fig:structure}, covering the Learnable Spectral Adversary, Siamese Spatiotemporal Feature Encoding, and shortcut-suppression optimization.

\subsection{Learnable Spectral Adversary}
\label{LSA}
Current detectors are highly sensitive to the amplitude statistics in specific temporal frequency bands, leading them to internalize spurious shortcut correlations that link spectral statistics directly to ground truth labels.
To mitigate this fragility, we propose the LSA component. 
It uses a generative network to synthesize the adversarial spectral deformation, encouraging the detector to learn forensic signatures that are invariant to unstable spectral statistics.
Following Eq.~(1), we define an environment transform $\mathcal{T}_e$ that perturbs only amplitude, with $P_m(\omega_k)$ fixed:
\begin{equation}
\widehat{X}_m(\omega_k)=\mathcal{T}_e(A_m(\omega_k))e^{jP_m(\omega_k)}.
\end{equation}

For window- and frame-rate-invariant parameterization, we use discrete normalized frequency bins 
$\Omega=\{\omega_k=k/T\mid k=0,\ldots,\lfloor T/2\rfloor\}$.
To find the most disruptive spectral deformations beyond hand-crafted heuristics, we parameterize the environment transform with a lightweight generator $G_{\phi}$.
Given a clip, we extract patch temporal profiles $\{x_{m}(t)\}_{m=1}^{M}$ from the facial ROI and apply a 1D temporal DFT to obtain the one-sided spectrum $X_m(\omega_k)$ as in Eq.~(1), evaluated on the discrete bins $\omega_k\in\Omega$.
Under \emph{phase consistency}, we fix $P_m(\omega_k)$ and perturb only the amplitude spectrum.
For each sample $i$ in a mini-batch of size $B$, the generator maps the per-clip min-max normalized amplitude $\breve{A}_{m,i}(\omega_k)$ (normalized over patches and frequencies) to a frequency-wise modulation field over DFT bins.
To enforce non-negativity of the amplitude spectrum and avoid implicit phase inversions, we use an exponential parameterization:
\begin{equation}
\widehat{A}_{m,i}(\omega_k)
=
A_{m,i}(\omega_k)\odot
\exp\!\Big(\alpha \cdot \tanh\big(G_{\phi}(\breve{A}_{m,i}(\omega_k))\big)\Big)
\label{eq:exp_amp}
\end{equation}
where $\odot$ denotes element-wise multiplication and $\alpha$ controls the perturbation strength.
The DC gain is fixed to one in implementation; pixel-wise RGB reconstruction is detailed in Appendix~\ref{app:rgb_reconstruction}.
This ensures $\widehat{A}_{m,i}(\omega_k)\geq 0$ for all $\omega_k\in\Omega$.
We then construct the perturbed spectrum 
$\widehat{X}_{m,i}(\omega_k)= \widehat{A}_{m,i}(\omega_k)e^{jP_{m,i}(\omega_k)}$ 
and reconstruct the adversarial temporal signal $\widehat{x}_{m,i}(t)$ via Inverse Discrete Fourier Transform (iDFT).
To ensure $\widehat{x}_{m,i}(t)\in\mathbb{R}$, we mirror the one-sided spectrum to negative frequencies to enforce Hermitian symmetry.
We formulate a minimax game between the generator and the detector.
The generator searches for adversarial spectral environments that maximally shift the detector's latent features, while the detector learns invariance to these shifts.
Let $f_{\theta}(\cdot)$ be the detector's feature extractor, and let $\{(X_{\mathrm{clean},i},y_i)\}_{i=1}^{B}$ denote a mini-batch of labeled videos.
We measure the distributional discrepancy between original and adversarial features using Maximum Mean Discrepancy (MMD) as:
\begin{equation}
\mathcal{D}_{\mathrm{MMD}}
=
{\mathrm{MMD}}
\Big(\{f_{\theta}(X_{\mathrm{clean},i})\}_{i=1}^{B}, \{f_{\theta}(X_{\mathrm{env},i})\}_{i=1}^{B}\Big).
\end{equation}
We define the effective modulation mask per sample, patch, and frequency as:
\begin{equation}
\mathcal{M}_{m,i}(\omega_k)
=
\frac{\widehat{A}_{m,i}(\omega_k)}{A_{m,i}(\omega_k)+\delta},
\label{eq:mask_def}
\end{equation}
where $\delta$ is a small constant for numerical stability.
The generator objective is:
\begin{equation}
\max_{\phi}\ \mathcal{L}_{\mathrm{gen}}
=
\frac{1}{B}\sum_{i=1}^{B}\mathcal{L}_{\mathrm{CE}}\big(p(X_{\mathrm{env},i}),y_i\big)
+
\gamma\,\mathcal{D}_{\mathrm{MMD}}
-
\frac{\lambda_{\mathrm{mask}}}{N}\big\| \mathcal{M} - \mathbf{1} \big\|_F^2,
\label{eq:gen_obj}
\end{equation}
where $\gamma\geq 0$ weights the feature-distribution shift term, $\lambda_{\mathrm{mask}}\geq 0$ regularizes the modulation mask toward identity to prevent trivial large perturbations, $\Omega$ denotes the discrete normalized DFT bins, $\mathcal{M} \in \mathbb{R}^{B \times M \times |\Omega|}$ is the modulation mask from Eq.~(6), $\mathbf{1}$ is an all-ones tensor of the same shape as $\mathcal{M}$, and $N = BM|\Omega|$ is the total number of elements.
The detector minimizes classification loss over original and adversarial samples:
\begin{equation}
\min_{\theta}\ \mathcal{L}_{\mathrm{det}}
=
\frac{1}{B}\sum_{i=1}^{B}
\Big(
\mathcal{L}_{\mathrm{CE}}\big(p(X_{\mathrm{clean},i}),y_i\big)
+
\mathcal{L}_{\mathrm{CE}}\big(p(X_{\mathrm{env},i}),y_i\big)
\Big).
\label{eq:det_obj}
\end{equation}

We optimize $\phi$ and $\theta$ alternately.
This adversarial training drives $G_{\phi}$ to expose the detector's reliance on spurious amplitude spectrum statistics, urging it to learn robust forensic representations that are invariant to spectral environments.

\subsection{Siamese Spatiotemporal Feature Encoding}

To capture long-range temporal dependencies, we adopt a Siamese design with a ViT-based~\cite{Dosovitskiy2020AnII} VideoMAE V2~\cite{Wang2023VideoMAEVS} backbone encoder, initialized from a checkpoint pretrained on large-scale video data.
Given an input clip $X \in \mathbb{R}^{T \times H \times W \times 3}$, VideoMAE V2 tokenizes $X$ into non-overlapping spatiotemporal tubelets.
Let $X_{\mathrm{clean}}$ denote the original clip and $X_{\mathrm{env}}$ the spectrally perturbed clip generated by LSA.
The tokens pass through Transformer blocks with joint space and time attention, allowing global interactions across spatiotemporal tokens and capturing temporal inconsistencies over the full clip.
We obtain clip-level representations $h_{\mathrm{clean}} \in \mathbb{R}^{D}$ and $h_{\mathrm{env}} \in \mathbb{R}^{D}$ by global average pooling over output tokens from the final layer of the clean and perturbed views, respectively.
During training, we use a weight-sharing Siamese strategy to map both views into a shared representation space.
The shared parameter constraint encourages the encoder to suppress environment-dependent spectral artifacts and retain robust cues that remain consistent across phase-preserving environments.
The dual-stream design is used only during training.
At inference, we only use $X_{\mathrm{clean}}$.

\subsection{Optimization for Shortcut Suppression}
To avoid VideoMAE's shortcut learning from spurious amplitude spectrum statistics and improve out-of-distribution (OOD) forensic generalization, we optimize the detector with a compound objective: (\emph{i}) symmetric prediction invariance at the prediction level and (\emph{ii}) spectral blindness at the representation level.

\noindent \textbf{Spectral Blindness.}
Prediction consistency alone does not prevent the encoder from encoding environment-specific amplitude statistics in latent space.
To suppress them, we enforce domain confusion between the clean and perturbed views.
We introduce a lightweight domain discriminator $q(\cdot)$ to predict a binary domain label
$d\in\{0,1\}$, where $d=0$ denotes the clean view and $d=1$ denotes the perturbed view.
With a Gradient Reversal Layer (GRL), we compute domain prediction:
\begin{equation}
\hat{d}_{\mathrm{clean}} = q(\mathrm{GRL}(h_{\mathrm{clean}})), \quad
\hat{d}_{\mathrm{env}}   = q(\mathrm{GRL}(h_{\mathrm{env}})).
\end{equation}
We then minimize the domain classification loss:
\begin{equation}
\mathcal{L}^{\mathrm{inv}}_{\mathrm{blind}}
=
\mathcal{L}_{\mathrm{CE}}(\hat{d}_{\mathrm{clean}}, 0)
+
\mathcal{L}_{\mathrm{CE}}(\hat{d}_{\mathrm{env}}, 1).
\label{eq:lblind}
\end{equation}
During backpropagation, the GRL reverses the sign of gradients into the encoder, equivalent to solving $\min_{q}\max_{\theta}\ \mathcal{L}^{\mathrm{inv}}_{\mathrm{blind}}$, driving the encoder to make $h_{\mathrm{clean}}$ and $h_{\mathrm{env}}$ indistinguishable so that controllable spectral cues are discarded and only phase-consistent evidence is retained.

\noindent \textbf{Symmetric Invariance.}
We align the predicted class distributions of $X_{\mathrm{clean}}$ and its spectrally perturbed counterpart $X_{\mathrm{env}}$ to encourage invariance to amplitude perturbations.
A prediction mismatch indicates sensitivity to the applied spectral perturbation.
We adopt a symmetric Kullback--Leibler (KL) divergence as a consistency regularizer:
\begin{equation}
\mathcal{L}^{\mathrm{inv}}_{\mathrm{sym}}
=
\frac{1}{2}\Big(
\mathcal{D}_{\mathrm{KL}}(p_{\mathrm{clean}}\Vert p_{\mathrm{env}})
+
\mathcal{D}_{\mathrm{KL}}(p_{\mathrm{env}}\Vert p_{\mathrm{clean}})
\Big),
\label{eq:linv}
\end{equation}
where $p_{\mathrm{clean}} = \mathrm{softmax}(g(h_{\mathrm{clean}}))$ and 
$p_{\mathrm{env}} = \mathrm{softmax}(g(h_{\mathrm{env}}))$ denote class distributions for the clean and perturbed views, and $g(\cdot)$ is the classification head.
Minimizing $\mathcal{L}^{\mathrm{inv}}_{\mathrm{sym}}$ promotes invariance to amplitude spectrum perturbations, reduces sensitivity to environment-dependent spectral artifacts.

Crucially, this prediction-level constraint remains essential even with adversarial representation learning. 
While the domain classification loss reduces attack-specific spectral information by aligning marginal distributions, adversarial training with a finite-capacity discriminator cannot guarantee strict feature alignment for each sample. 
Without $\mathcal{L}^{\mathrm{inv}}_{\mathrm{sym}}$, subtle residual feature discrepancies can be amplified by the classification head and lead to unpredictable changes in predictions for individual samples. 
$\mathcal{L}^{\mathrm{inv}}_{\mathrm{sym}}$ prevents this vulnerability at the decision level by serving as an instance-level anchor.

\noindent \textbf{Total Objective.}
% \subsubsection{Total Objective}
Let $y$ be the manipulation label and let $d\in\{0,1\}$ denote the domain label,
where $d=0$ corresponds to the clean view $X_{\mathrm{clean}}$ and $d=1$ corresponds to the spectrally perturbed view $X_{\mathrm{env}}$ generated by LSA.
The final training objective is defined as follows:
\begin{equation}
\mathcal{L}_{\mathrm{total}}
=
\mathcal{L}_{\mathrm{det}}
+
\lambda_{\mathrm{sym}}\mathcal{L}^{\mathrm{inv}}_{\mathrm{sym}}
+
\lambda_{\mathrm{blind}}\mathcal{L}^{\mathrm{inv}}_{\mathrm{blind}},
\label{eq:ltotal}
\end{equation}
where $\lambda_{\mathrm{sym}}$ and $\lambda_{\mathrm{blind}}$ are non-negative hyperparameters that balance symmetric prediction invariance, spectral blindness, and the primary detection objective.

\section{Experiment}

\subsection{Experiment Setting}

\textbf{Dataset.}
We evaluate SpInShield on the following datasets:
Celeb-DF-v2 (CDF-v2) \cite{cdfv2}, DeepFakeDetection (DFD) dataset \cite{dfd}, Deeperforensics (DFo) dataset \cite{dfo} and WildDeepFake (WDF) dataset \cite{wdf} using a model trained on the FF++ \cite{FF++}. 
To further assess the effects of GAN and Diffusion based generators, we add two evaluation setups using DiffSwap \cite{diffswap} and DaGAN \cite{DaGAN} on the FFHQ dataset \cite{ffhq}, each comprising 1,000 real samples and 1,000 fake samples. 

\noindent\textbf{Data Preprocessing.}
Following recent practices \cite{stylelatentflow,dfdfcg}, we utilize Dlib \cite{dlib} to extract facial landmarks for alignment, and Retinaface \cite{retinaface} for face cropping. 
To mitigate spurious temporal artifacts caused by inter-frame misalignment, we enforce strict spatial consistency by fixing the bounding box coordinates to the initial frame.
For temporal sampling, we randomly extract non-overlapping clips of 2 to 4 seconds from each video. 
From these clips, we extract a sequence of $T=16$ consecutive frames to maintain the continuity of the frequency spectrum. 
Finally, these frames undergo a comprehensive video-level augmentation pipeline to simulate real-world variations.
For more implementation details, please refer to Appendix~\ref{app:protocol}.

\noindent\textbf{Baselines.}
For a comprehensive evaluation, we compare SpInShield with the following advanced and representative baselines, which are categorized into:
\textit{Frame-level methods}: SLADD~\cite{SLADD}, SBI~\cite{SBI}, UCF~\cite{UCF}, IID~\cite{IID}, LSDA~\cite{LSDA}, ProDet~\cite{ProDet}, and CDFA~\cite{CDFA};
\textit{Video-level methods}: TALL~\cite{TALL}, SLF~\cite{stylelatentflow}, NACO~\cite{NACO}, AFW~\cite{AFW}, AltFreezing~\cite{altfreezing}, FTCN~\cite{FTCN}, PTFD~\cite{pixelwise}, FakeSTormer~\cite{FakeSTormer}, and DFD-FCG~\cite{dfdfcg}.
All baseline numbers reported in this paper are obtained from our own evaluations. 
We use the corresponding implementations and model-specific preprocessing, and evaluate each method under the protocol specified for the experiment.

\begin{table*}[t]
\renewcommand{\arraystretch}{0.9}
\centering
\caption{Cross-dataset evaluation on FF++ (c23). \textbf{Avg.} excludes in-domain results. DiF.\ (DiffSwap) and DaG.\ (DaGAN) denote evaluation sets constructed using FFHQ and the respective generators, each containing 1,000 real and 1,000 fake samples. $^*$ indicates statistically significant improvements over the strongest baseline ($p < 0.05$, two-tailed t-test) on the average metric.}
\label{tab:benchmark_results_compact}
\setlength{\tabcolsep}{3.8pt}
\scriptsize
\begin{tabular}{l *{17}{c} c}
\toprule
& \rotatebox{90}{SLADD\,}
& \rotatebox{90}{SBI\,}
& \rotatebox{90}{UCF\,}
& \rotatebox{90}{IID\,}
& \rotatebox{90}{LSDA\,}
& \rotatebox{90}{ProDet\,}
& \rotatebox{90}{CDFA\,}
& \rotatebox{90}{TALL\,}
& \rotatebox{90}{SLF\,}
& \rotatebox{90}{NACO\,}
& \rotatebox{90}{AFW\,}
& \rotatebox{90}{AltFre\,}
& \rotatebox{90}{FTCN\,}
& \rotatebox{90}{PTFD\,}
& \rotatebox{90}{FakeST\,}
& \rotatebox{90}{DFD-FCG\,}
& \cellcolor{blue!5}\rotatebox{90}{\textbf{Ours}\,} \\
\midrule
\textbf{FF++} \scriptsize\textit{(In-D.)}
& .984 & .997 & .995 & .993 & .990 & .992 & .993 & \textbf{.998} & .984 & .997 & .980 & .995 & .997 & .984 & .995 & .993
& \cellcolor{blue!5}.996 \\
\midrule
CDFv2
& .837 & .886 & .837 & .838 & .875 & .926 & .938 & .871 & .890 & .895 & .824 & .890 & .869 & .914 & .924 & \textbf{.949}
& \cellcolor{blue!5}.948 \\
DFD
& .904 & .827 & .867 & .939 & .881 & .901 & .954 & .892 & .961 & .970 & .860 & .902 & .879 & .920 & \textbf{.985} & .928
& \cellcolor{blue!5}.981 \\
DFo
& .800 & .899 & .808 & .808 & .768 & .879 & .973 & \textbf{.996} & .990 & .995 & .819 & .990 & .988 & .954 & .960 & .994
& \cellcolor{blue!5}.987 \\
WDF
& .690 & .703 & .774 & .666 & .797 & .781 & .796 & .680 & .741 & .686 & .693 & .773 & .791 & .741 & .766 & .875
& \cellcolor{blue!5}\textbf{.889} \\
DiffSwap
& .691 & .704 & .784 & .729 & .703 & .701 & .750 & .689 & .748 & .722 & .692 & .745 & .752 & .820 & .845 & .872
& \cellcolor{blue!5}\textbf{.894} \\
DaGAN
& .730 & .743 & .722 & .683 & .793 & .693 & .728 & .700 & .831 & .741 & .753 & .802 & .783 & .874 & .820 & .884
& \cellcolor{blue!5}\textbf{.897} \\
\midrule
\textbf{Avg.}
& .775 & .794 & .799 & .777 & .803 & .814 & .857 & .805 & .860 & .835 & .774 & .850 & .844 & .871 & .883 & .917
& \cellcolor{blue!5}\textbf{.933$^*$} \\
\bottomrule
\end{tabular}
\vspace{-4mm}
\end{table*}

\subsection{Non-Adversarial Evaluation}
This section validates SpInShield's robustness against standard deepfake generators as a baseline. 
In \autoref{tab:benchmark_results_compact}, SpInShield obtains state-of-the-art (SOTA) results in cross-domain evaluation, obtaining an average AUC of 93.3\%. 
This surpasses the second-best competitor (DFD-FCG) by a margin of 1.6\%.
This generalization stems from our temporal spectral synthesis design.
By explicitly exposing the model to diverse simulated spectral environments in training, we encourage the encoder to discount spurious amplitude statistics that vary across datasets and recording pipelines. 
Furthermore, the shortcut suppression optimization urges the model to effectively learn phase spectrum information.
% CR 3 4
A matched VideoMAE V2 baseline further indicates that the clean generalization gains cannot be attributed to backbone pretraining alone (Appendix~\ref{app:matched_backbone}).
Additional evaluations on FFIW and a DF40 talking-face subset show comparable clean performance to DFD-FCG and higher robustness under spectral perturbations (Appendix~\ref{app:extended_benchmarks}).

\begin{figure*}[t]
    \centering
          \includegraphics[width=0.95\linewidth]{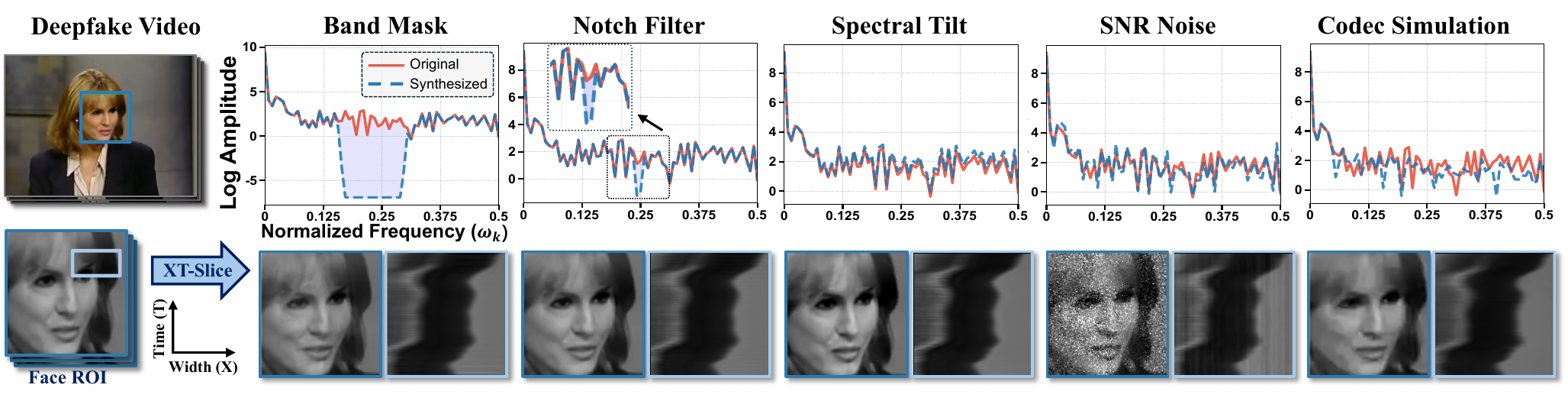}
    \caption{Visualization of amplitude spectral attack synthesis. Facial XT-slices are extracted from the input video to generate diverse attacks. The top row compares the log-amplitude spectra of the original (red solid) and synthesized (blue dashed) signals over normalized DFT-bin frequencies $\omega_k$. The bottom row shows the corresponding visual effects. Perturbations are amplified for visualization only.}
    % \label{fig:defined_env}
    \label{fig:defined_env}
    \vspace{-4mm}
\end{figure*}

\subsection{Resilience to Temporal Spectral Attack}
\label{ar}

\noindent \textbf{Robustness Against Synthetic Spectral Attacks.}
To evaluate the generalizability against unseen amplitude spectral conditions, we define a set of spectral attacks $\mathcal{E}$ as benchmarks. 
We construct these attacks following~ Section \ref{LSA} with specific, fixed transformations $\mathcal{T}_e$ to the amplitude spectrum $A(\omega_k)$, and reconstruct the evaluation samples via iDFT.
Each attack $e\in\mathcal{E}$ is categorized into Type-I: de-correlation via masking/reshaping, and Type-II: robustness via channel/platform distortions:
\textit{Random Band Masking:}
    We suppress $B\sim\mathrm{Unif}\{1,2,3\}$ randomly placed contiguous DFT-bin bands with Tukey-tapered masks, where each band covers $K_{\mathrm{band}}\sim\mathrm{Unif}\{1,2,3,4\}$ bins;
\textit{Narrow-band Notch:}
    We attenuate localized DFT-bin regions using a smooth cosine-tapered notch mask. The center bin is sampled uniformly from non-DC and non-Nyquist bins, with a total width of $K_{\mathrm{notch}}\sim\mathrm{Unif}\{1,2\}$ bins;
\textit{Spectral Tilt:}
    We re-weight each DFT bin in the log-amplitude domain as $\log \widehat{A}(\omega_k)=\log(A(\omega_k)+\epsilon_0)+\beta_1\omega_k+\beta_2\omega_k^2$, with $\beta_1,\beta_2\sim U(-1.5,1.5)$;
\textit{SNR-based Noise:}
    We inject per-bin Gaussian noise in the log-amplitude domain: $\log \widehat{A}(\omega_k)=\log(A(\omega_k)+\epsilon_0)+\eta(\omega_k)$, where $\eta(\omega_k)\sim\mathcal{N}(0,\sigma^2)$.
\autoref{fig:defined_env} presents the amplitude spectral deviations of each attack.

\autoref{tab:stress_test_final_compact_incomplete} reports AUC under four amplitude spectral attacks across six datasets.
SpInShield consistently outperforms across all attacks and datasets, with average AUC scores of 89.35, 90.47, 88.47, and 87.43, respectively.
In contrast, baselines suffer substantial degradation under these attacks, with most drop below AUC of 70\%, while the VideoMAE baselines approach random-level performance under SNR-based Noise (54.60\%).
Notably, SpInShield maintains an average absolute improvement of over 20 percentage points in AUC compared with the strongest baselines across all attack types, showing that the invariance learned through LSA and Shortcut Suppression Optimization generalizes effectively to unseen, hand-crafted spectral attacks beyond those encountered during training.
We assess the perceptual severity of the four spectral attacks using frame-level quality and optical-flow consistency metrics. The measurements suggest limited appearance and motion distortion under the evaluated budgets (Appendix~\ref{app:perceptual_distortion}).
Retraining on temporally smoothed FF++ improves baseline robustness but leaves a substantial gap to SpInShield (Appendix~\ref{app:temporal_smoothing}).

% \vspace{-4pt}
\noindent \textbf{Robustness in Real World Deployment Scenarios.}
We further evaluate SpInShield under real world video processing pipelines to assess its practical robustness under non differentiable post processing operations.
To evaluate generalization in practical deployment, we assess our method under four real world video post processing pipelines: platform round trips, multi stage codec chains, heavy compression, and cross platform cascades. 
As \autoref{tab:real_world_pipeline_results_incomplete} shows, SpInShield consistently achieves the highest average AUC across all settings, maintaining performance between 83.28\% and 89.92\%. While baseline methods suffer substantial performance drops under severe compression, the phase consistent forensic signals learned by SpInShield remain stable and outperform all competitors. 

\begin{table*}[t] 
    \centering 
    \caption{Robustness evaluation under amplitude spectrum attacks. $^*$ indicates statistically significant improvements over the strongest baseline via a two-tailed t-test ($p < 0.05$) on the average metric.}
    \label{tab:stress_test_final_compact_incomplete}
    \resizebox{\textwidth}{!}{
        \scriptsize
        \renewcommand{\arraystretch}{0.85} 
        \setlength{\tabcolsep}{4pt}    
        \begin{tabular}{l ccccccc @{\hspace{5pt}} ccccccc} 
        \toprule
        \textbf{Method} & \textbf{CDFv2} & \textbf{DFD} & \textbf{DFo} & \textbf{WDF} & \textbf{DiF.} & \textbf{DaG.} & \textbf{Avg.} & \textbf{CDFv2} & \textbf{DFD} & \textbf{DFo} & \textbf{WDF} & \textbf{DiF.} & \textbf{DaG.} & \textbf{Avg.} \\
        \midrule
        
        & \multicolumn{7}{c}{\textbf{(a) Random Band Mask}} & \multicolumn{7}{c}{\textbf{(b) Narrow-band Notch}} \\
        \cmidrule(r{5pt}){2-8} \cmidrule{9-15}
        
        FakeST      & 70.21 & 73.35 & 75.46 & 58.52 & 58.39 & 56.65 & 65.43 & 72.55 & 76.68 & 78.53 & 60.71 & 60.45 & 59.80 & 68.12 \\
        AltFre      & 65.23 & 73.55 & 76.22 & 60.19 & 62.50 & 68.59 & 67.71 & 68.48 & 76.55 & 79.23 & 62.82 & 64.80 & 71.23 & 70.52 \\
        PTFD        & 66.18 & 71.21 & 72.85 & 59.52 & 68.54 & 74.50 & 68.80 & 69.24 & 74.53 & 75.63 & 61.54 & 70.24 & 76.80 & 71.33 \\
        TALL        & 63.58 & 70.85 & 77.13 & 58.44 & 58.24 & 59.53 & 64.63 & 65.85 & 73.45 & 79.59 & 59.65 & 60.49 & 61.86 & 66.81 \\
        SLF         & 67.21 & 75.42 & 76.81 & 61.22 & 63.54 & 70.20 & 69.07 & 69.52 & 77.81 & 79.16 & 63.42 & 65.80 & 72.50 & 71.37 \\
        NACO        & 66.85 & 76.16 & 78.44 & 59.80 & 61.50 & 62.89 & 67.61 & 68.85 & 78.57 & 80.25 & 61.24 & 63.82 & 65.20 & 69.66 \\
        AFW         & 60.51 & 66.83 & 64.23 & 55.42 & 57.80 & 62.56 & 61.23 & 62.46 & 68.55 & 66.73 & 57.25 & 59.54 & 64.89 & 63.24 \\
        DFD-FCG     & 72.28 & 75.56 & 79.41 & 59.25 & 59.48 & 58.38 & 67.39 & 74.35 & 78.42 & 82.34 & 62.24 & 61.28 & 60.46 & 69.85 \\
        FTCN        & 64.52 & 68.25 & 75.53 & 58.89 & 64.20 & 66.58 & 66.33 & 66.55 & 71.41 & 78.20 & 60.52 & 66.54 & 68.87 & 68.68 \\
        VideoMAE(B) & 60.15 & 74.25 & 55.42 & 54.80 & 54.08 & 56.51 & 59.20 & 61.83 & 76.51 & 56.16 & 56.06 & 55.81 & 58.23 & 60.77 \\
        \textbf{SpInShield} 
        & \textbf{91.82} & \textbf{90.88} & \textbf{94.79} & \textbf{86.91} & \textbf{86.47} & \textbf{85.23} & \textbf{89.35 $^*$} 
        & \textbf{92.61} & \textbf{91.98} & \textbf{96.12} & \textbf{87.91} & \textbf{87.88} & \textbf{86.32} & \textbf{90.47 $^*$} \\
        
        \midrule 
        & \multicolumn{7}{c}{\textbf{(c) Spectral Tilt}} & \multicolumn{7}{c}{\textbf{(d) SNR-based Noise}} \\
        \cmidrule(r{5pt}){2-8} \cmidrule{9-15}

        FakeST      & 65.24 & 70.41 & 72.48 & 55.26 & 54.29 & 53.53 & 61.87 & 62.38 & 66.45 & 69.36 & 51.48 & 50.39 & 50.50 & 58.43 \\
        AltFre      & 62.44 & 70.58 & 75.65 & 58.24 & 59.61 & 64.50 & 65.17 & 60.15 & 68.27 & 72.56 & 55.43 & 56.40 & 61.24 & 62.34 \\
        PTFD        & 63.85 & 69.25 & 71.52 & 57.49 & 66.44 & 71.24 & 66.63 & 61.22 & 66.51 & 68.40 & 54.52 & 61.53 & \textbf{66.40} & 63.10 \\
        TALL        & 61.55 & 68.48 & 74.23 & 56.52 & 55.62 & 56.48 & 62.15 & 58.55 & 65.42 & 71.22 & 53.25 & 52.11 & 53.80 & 59.06 \\
        SLF         & 64.55 & 72.59 & 75.81 & 59.25 & 60.23 & 67.40 & 66.64 & 61.88 & 69.55 & 72.44 & 56.56 & 57.50 & 62.81 & 63.46 \\
        NACO        & 64.20 & 74.11 & 76.58 & 58.83 & 58.43 & 59.52 & 65.28 & 62.15 & 71.21 & 74.52 & 55.85 & 55.29 & 56.40 & 62.57 \\
        AFW         & 58.52 & 64.21 & 62.55 & 53.42 & 54.26 & 58.68 & 58.61 & 56.45 & 61.53 & 60.23 & 51.53 & 51.59 & 55.20 & 56.09 \\
        DFD-FCG     & 68.52 & 72.63 & 75.71 & 56.78 & 56.64 & 54.62 & 64.15 & 64.75 & 68.58 & 72.69 & 53.57 & 52.48 & 51.65 & 60.62 \\
        FTCN        & 62.52 & 67.51 & 73.45 & 57.82 & 61.53 & 63.40 & 64.37 & 60.40 & 65.25 & 70.57 & 55.24 & 58.20 & 60.59 & 61.71 \\
        VideoMAE(B) & 57.43 & 70.12 & 53.81 & 53.21 & 52.37 & 53.51 & 56.74 & 55.20 & 66.80 & 52.13 & 51.46 & 50.81 & 51.20 & 54.60 \\
        \textbf{SpInShield} 
        & \textbf{90.52} & \textbf{90.18} & \textbf{94.81} & \textbf{86.11} & \textbf{85.28} & \textbf{83.92} & \textbf{88.47 $^*$} 
        & \textbf{89.82} & \textbf{89.48} & \textbf{93.11} & \textbf{85.41} & \textbf{84.18} & \textbf{82.58} & \textbf{87.43 $^*$} \\
        \bottomrule
        \end{tabular}
    }
\end{table*}

\begin{table*}[t]
    \centering
    \caption{Robustness under real-world video processing pipelines (AUC \%).
    $^*$ indicates statistically significant improvements over the strongest baseline via a two-tailed t-test ($p < 0.05$).}
    \label{tab:real_world_pipeline_results_incomplete}
    \renewcommand{\arraystretch}{0.9}
    \setlength{\tabcolsep}{3pt}
    \scriptsize
    \begin{tabular}{l cccccccccc c}
    \toprule
    \textbf{Pipeline}
    & \textbf{FakeST} & \textbf{AltFre} & \textbf{PTFD} & \textbf{TALL} & \textbf{SLF} & \textbf{NACO} & \textbf{AFW} & \textbf{DFD-FCG} & \textbf{FTCN} & \textbf{VideoMAE(B)} & \textbf{SpInShield} \\
    \midrule
    Black-Box Platform  & 82.38 & 80.63 & 82.97 & 78.32 & 83.11 & 81.23 & 73.73 & 76.08 & 81.02 & 70.94 & \textbf{89.92}$^*$ \\
    Multi-stage Codec   & 77.53 & 76.37 & 78.72 & 73.99 & 78.58 & 77.02 & 69.16 & 75.27 & 76.54 & 65.64 & \textbf{87.53}$^*$ \\
    Heavy Codec         & 71.56 & 70.66 & 72.81 & 67.52 & 72.95 & 70.89 & 63.71 & 75.38 & 71.17 & 58.80 & \textbf{83.28}$^*$ \\
    Cross-Platform      & 75.53 & 73.71 & 76.27 & 71.47 & 76.31 & 74.43 & 66.84 & 75.61 & 74.34 & 64.24 & \textbf{84.44}$^*$ \\
    \bottomrule
    \end{tabular}
\end{table*}

\begin{wraptable}{r}[-10pt]{0.4\textwidth}
    \centering
    \vspace{-10mm}
    \caption{Adaptive white-box evaluation. VMAE denotes the plain VideoMAE.}
    \label{tab:adaptive_attack_compact}
    \scriptsize
    \renewcommand{\arraystretch}{0.7}
    \setlength{\tabcolsep}{1.5pt}
    \begin{tabular}{l ccccc}
    \toprule
    & \textbf{FakeST} & \textbf{SLF} & \textbf{DFD-FCG} & \textbf{VMAE} & \textbf{Ours} \\
    \midrule
    CDFv2    & 54.27 & 55.74 & 60.43 & 47.83 & \textbf{67.74} \\
    DFD      & 58.83 & 60.18 & 65.78 & 53.62 & \textbf{66.42} \\
    DFo      & 60.14 & 62.45 & 64.92 & 45.91 & \textbf{72.86} \\
    WDF      & 47.62 & 49.61 & 53.47 & 43.27 & \textbf{61.83} \\
    DiffSwap & 48.91 & 51.36 & 55.86 & 44.58 & \textbf{63.27} \\
    DaGAN    & 47.36 & 56.27 & 56.21 & 45.18 & \textbf{63.91} \\
    \midrule
    Avg.     & 52.86 & 55.94 & 59.45 & 46.73 & \textbf{66.01} \\
    \bottomrule
    \end{tabular}
    \vspace{-10pt}
\end{wraptable}

\vspace{-1mm}
\subsection{Adaptive Evaluation against SpInShield}
\label{adaptive_attack_main}

We conduct an adaptive white box evaluation where attackers optimize video specific amplitude modulation masks. As \autoref{tab:adaptive_attack_compact} shows, Baseline models degrade significantly, but SpInShield retains the highest average AUC of 66.01\%.
We additionally evaluate pixel-space PGD attacks not used during training and observe higher AUC than DFD-FCG at both tested perturbation budgets (Appendix~\ref{app:pixel_pgd}).

\subsection{Phase-Only Evaluation}
To validate whether SpInShield uses phase information more effectively, we evaluate the performance in a setting where all amplitude spectrum information is removed.
\autoref{tab:combined} (a) reports the performance.
Existing baselines suffer significant performance degradation with phase spectrum information only, exhibiting AUC drops from 24.9\% to 29.6\%.
This performance gap demonstrates their heavy reliance on amplitude spectrum information.
Conversely, SpInShield maintains an AUC of 79.6\% with the minimum performance gap of 13.7\% among all evaluated methods.
These results suggest that SpInShield effectively uses phase information for reliable deepfake detection.

\begin{table*}[t]
\centering
\caption{(a) Robustness under full information and phase-only input. (b) Component contributions on in-domain performance and OOD robustness.}
\label{tab:combined}
\vspace{-2mm}
\begin{minipage}[t]{0.32\textwidth}
\renewcommand{\arraystretch}{1.0}
\centering
\scriptsize
\subcaption{Full info.\ vs.\ phase-only.}\label{tab:phase_vertical_final}
\vspace{-2mm}
\renewcommand{\arraystretch}{0.81}
\setlength{\tabcolsep}{7pt}
\begin{tabular}{l ccc}
\toprule
& \scriptsize\textbf{Full} & \scriptsize\textbf{Phase} & \scriptsize$\boldsymbol{\Delta}$ \\
\midrule
\scriptsize FakeST.   & \scriptsize 88.3 & \scriptsize 60.2 & \scriptsize -28.1 \\
\scriptsize AltFre.   & \scriptsize 85.0 & \scriptsize 58.7 & \scriptsize -26.3 \\
\scriptsize PTFD      & \scriptsize 87.1 & \scriptsize 57.5 & \scriptsize -29.6 \\
\scriptsize TALL      & \scriptsize 80.5 & \scriptsize 55.6 & \scriptsize -24.9 \\
\scriptsize SLF       & \scriptsize 86.0 & \scriptsize 59.4 & \scriptsize -26.6 \\
\scriptsize NACO      & \scriptsize 83.5 & \scriptsize 58.2 & \scriptsize -25.3 \\
\scriptsize AFW       & \scriptsize 77.4 & \scriptsize 52.1 & \scriptsize -25.3 \\
\scriptsize DFD-FCG   & \scriptsize 91.7 & \scriptsize 64.5 & \scriptsize -27.2 \\
\scriptsize FTCN      & \scriptsize 84.4 & \scriptsize 57.8 & \scriptsize -26.6 \\
\midrule
\scriptsize\textbf{Ours} & \scriptsize\textbf{93.3} & \scriptsize\textbf{79.6} & \scriptsize\textbf{-13.7} \\
\bottomrule
\end{tabular}%
\end{minipage}%
\hfill
\begin{minipage}[t]{0.65\textwidth}
\centering
\subcaption{Ablation study.}\label{tab:ablation}
\vspace{-2mm}
\renewcommand{\arraystretch}{0.67}
\setlength{\tabcolsep}{4pt}
\begin{tabular}{l c *{7}{c}}
\toprule
\multirow{2.5}{*}{\scriptsize\textbf{Method}}
& \scriptsize\textbf{In-Dom.}
& \multicolumn{7}{c}{\scriptsize\textbf{OOD}} \\
\cmidrule(lr){2-2} \cmidrule(lr){3-9}
& \rotatebox{90}{\scriptsize FF++\,}
& \rotatebox{90}{\scriptsize CDF-v2\,}
& \rotatebox{90}{\scriptsize DFD\,}
& \rotatebox{90}{\scriptsize DFo\,}
& \rotatebox{90}{\scriptsize WDF\,}
& \rotatebox{90}{\scriptsize DS\,}
& \rotatebox{90}{\scriptsize DaGAN\,}
& \rotatebox{90}{\scriptsize Avg.\,} \\
\midrule
\scriptsize VideoMAE(B)         & \scriptsize 86.89 & \scriptsize 58.66 & \scriptsize 71.93 & \scriptsize 54.38 & \scriptsize 53.88 & \scriptsize 53.27 & \scriptsize 54.86 & \scriptsize 57.83 \\
\scriptsize w/ Naive            & \scriptsize 94.81 & \scriptsize 62.19 & \scriptsize 65.98 & \scriptsize 64.32 & \scriptsize 62.68 & \scriptsize 59.29 & \scriptsize 64.87 & \scriptsize 63.22 \\
\scriptsize w/o $\mathcal{L}^{\mathrm{inv}}_{\mathrm{sym}}$
                    & \scriptsize 96.10 & \scriptsize 80.35 & \scriptsize 88.65 & \scriptsize 81.45 & \scriptsize 74.60 & \scriptsize 73.80 & \scriptsize 80.51 & \scriptsize 79.89 \\
\scriptsize w/o $\mathcal{L}^{\mathrm{inv}}_{\mathrm{blind}}$
                    & \scriptsize 97.23 & \scriptsize 85.65 & \scriptsize 90.20 & \scriptsize 86.80 & \scriptsize 80.50 & \scriptsize 71.54 & \scriptsize 74.30 & \scriptsize 81.50 \\
\scriptsize\textbf{Complete}   & \scriptsize\textbf{99.60} & \scriptsize\textbf{91.26} & \scriptsize\textbf{90.60} & \scriptsize\textbf{94.75} & \scriptsize\textbf{86.63} & \scriptsize\textbf{85.93} & \scriptsize\textbf{84.45} & \scriptsize\textbf{88.94} \\
\bottomrule
\end{tabular}%

\end{minipage}
\end{table*}

\begin{figure}[t]
  \centering
   \includegraphics[width=1\textwidth]{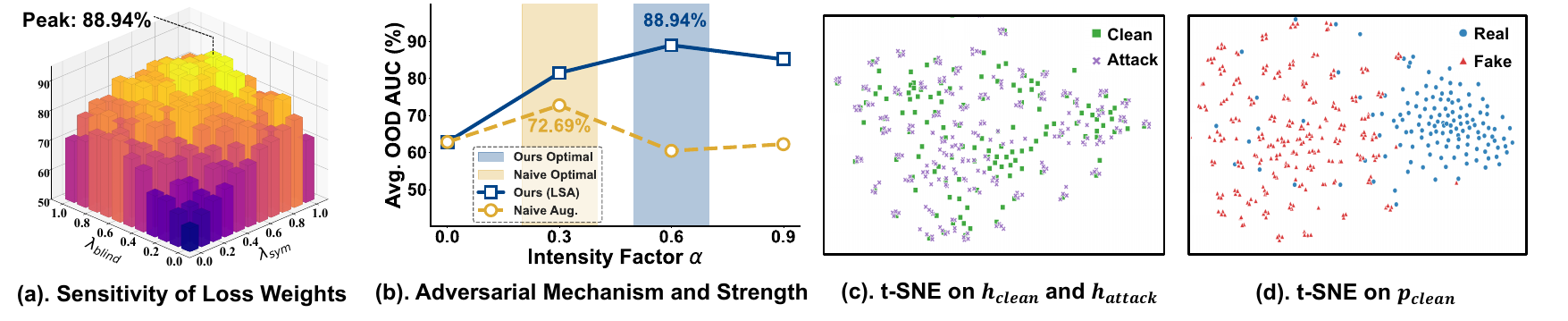}
  \caption{Quantitative and qualitative evaluation: (a) Joint impact of hyperparameters $\lambda_{sym}$ and $\lambda_{blind}$ on OOD generalization performance. (b) Impact of the amplitude spectral attack intensity $\alpha$ on the OOD generalization capability. (c) Feature space distribution visualization between clean and perturbed samples. (d) Decision space clustering visualization of real and fake video samples.}
  \label{fig:tsne} 
  \vspace{-2mm}
\end{figure}

\subsection{Ablation Study}
To isolate each component's contribution within SpInShield, we conduct an ablation study on FF++ for in-domain evaluation, and report OOD performance as average AUC under four amplitude spectral attacks.
In ~\autoref{tab:combined} (b), we add components on top of the VideoMAE backbone: Naive Spectral Augmentation with random Gaussian noise on amplitude spectrum (w/o LSA), $\mathcal{L}^{\mathrm{inv}}_{\mathrm{sym}}$, and $\mathcal{L}^{\mathrm{inv}}_\mathrm{blind}$.
The baseline shows in-domain performance with an AUC of 86.89\% but degrades severely in OOD settings with 57.83\% on average, confirming the amplitude shortcut problem.
Naive spectral augmentation improves marginally with +5.39\%, as random noise fails to synthesize targeted adversarial deformations that expose amplitude shortcuts effectively.
% camera ready 2
Removing $\mathcal{L}^{\mathrm{inv}}_{\mathrm{sym}}$ or $\mathcal{L}^{\mathrm{inv}}_{\mathrm{blind}}$ individually yields further gains but leaves a substantial gap, showing that prediction-level consistency and feature-level domain confusion are complementary.

A matched-objective comparison with fixed spectral augmentation further supports the benefit of the learnable adversary under comparable perturbation strength.

\noindent\textbf{Matched-objective comparison.}
To isolate the contribution of the Learnable Spectral Adversary (LSA), we compare SpInShield with a Fixed-SA variant.
Fixed-SA retains the same Siamese training pipeline, classification objective, prediction-consistency loss $\mathcal{L}^{\mathrm{inv}}_{\mathrm{sym}}$, and domain-confusion loss $\mathcal{L}^{\mathrm{inv}}_{\mathrm{blind}}$.
The backbone, optimizer, learning rate, training schedule, data split, and evaluation protocol are also unchanged.
The only difference is that LSA is replaced by fixed spectral augmentation.

To construct Fixed-SA, we apply the converged LSA from the full SpInShield model to the validation split and collect the mean and standard deviation of its log-amplitude masks for each temporal-frequency bin.
These statistics parameterize a fixed mask-sampling distribution during Fixed-SA training.
No test data are used to calibrate this distribution.
Both variants use the same amplitude-only modulation, spectral reconstruction, and clipping procedure.
The measured mean perturbation-strength ratio of Fixed-SA to LSA is 0.97, indicating comparable perturbation strength.

\begin{table}[htbp]
    \centering
    \captionsetup{position=top,skip=7pt}
    \caption{Matched-objective ablation of LSA. Both variants use the same losses and training settings with comparable perturbation strength. Results are average OOD AUC under the four spectral attacks.}
    \label{tab:fixed_sa}
    \small
    \begin{tabular}{lc}
        \toprule
        Variant & OOD Avg. AUC (\%) \\
        \midrule
        Fixed-SA & 78.37 \\
        SpInShield & \textbf{88.94} \\
        \bottomrule
    \end{tabular}
\end{table}

As shown in Table~\ref{tab:fixed_sa}, SpInShield outperforms Fixed-SA by 10.57 AUC percentage points.
Unlike Fixed-SA, whose sampling distribution remains unchanged throughout training, LSA is jointly optimized against the evolving detector.
This matched-objective comparison supports the benefit of adaptive spectral environment generation beyond fixed augmentation with comparable perturbation strength.

\vspace{-1mm}
\subsection{Sensitivity Analysis}
\vspace{-1mm}
\label{sec:loss_weight}
\autoref{fig:tsne} (a) demonstrates that SpInShield performs best when $\lambda_{\mathrm{blind}}=0.7$ and $\lambda_{\mathrm{sym}}=0.9$.
It illustrates that optimizing both attack invariance and spectral blindness can suppress amplitude spectrum shortcuts under a proper loss weight setup as insufficient adversarial weight setup fails to eliminate amplitude spectrum shortcuts.
\autoref{fig:tsne} (b) shows that SpInShield obtains a peak AUC of 88.94\% when $\alpha=0.6$, illustrating the impact of the perturbation intensity. 
Insufficient perturbation fails to suppress amplitude shortcuts, whereas excessive intensity destroys forensic cues and degrades performance. 
Moreover, LSA shows stronger performance compared to naive augmentation.

\vspace{-1mm}
\subsection{Latent Space Visualization}
\vspace{-1mm}
We visualize feature distributions and validate SpInShield's spectrum shortcut suppression via t-SNE.
\autoref{fig:tsne} (c) plots the encoder representations $h_{\mathrm{clean}}$ and $h_{\mathrm{env}}$.
The overlap between these two distributions shows the encoder is amplitude-blind, mapping both clean and perturbed inputs into a shared invariant latent space.
\autoref{fig:tsne} (d) illustrates that $h_{\mathrm{clean}}$ forms distinct clusters for real and fake samples in the decision space.
This clear separation demonstrates that the model retains strong discriminative capability for forgery classification after suppressing amplitude spectrum shortcuts.

\vspace{-1mm}
\section{Conclusion}
\vspace{-1mm}
This work addresses the limitations of existing spatiotemporal deepfake detectors in relying on shortcut learning with fragile temporal spectrum cues. 
We propose SpInShield framework, building a learnable spectral adversary component to synthesize severe amplitude spectrum deformations that expose latent vulnerabilities. 
Building upon LSA, we introduce a dual-objective shortcut suppression strategy: spectral blindness removes amplitude-specific information from the latent space, while symmetric invariance anchors prediction consistency at the decision level, collectively compelling the encoder to prioritize phase-consistent forensic cues.
SpInShield achieves advanced performance under simulated temporal spectral attacks, demonstrating its effectiveness.

\begin{ack}
The authors received no third-party funding or support for the work reported in this paper. The authors further declare that they have no competing interests, financial or otherwise, that could reasonably be perceived to have influenced the conduct of the research, the interpretation of its findings, or the presentation of its conclusions.
\end{ack}

\bibliographystyle{plainnat}
\bibliography{references}

@article{Mirsky2020TheCA,
  title={The Creation and Detection of Deepfakes},
  author={Yisroel Mirsky and Wenke Lee},
  journal={ACM Computing Surveys (CSUR)},
  year={2020},
  volume={54},
  pages={1 - 41},
}

@article{Geirhos2020ShortcutLI,
  title={Shortcut learning in deep neural networks},
  author={Geirhos, Robert and Jacobsen, J{\"o}rn-Henrik and Michaelis, Claudio and Zemel, Richard and Brendel, Wieland and Bethge, Matthias and Wichmann, Felix A},
  journal={Nature Machine Intelligence},
  volume={2},
  number={11},
  pages={665--673},
  year={2020},
  publisher={Nature Publishing Group UK, London}
}

@inproceedings{Wang2023VideoMAEVS,
  title={Videomae v2: Scaling video masked autoencoders with dual masking},
  author={Wang, Limin and Huang, Bingkun and Zhao, Zhiyu and Tong, Zhan and He, Yinan and Wang, Yi and Wang, Yali and Qiao, Yu},
  booktitle={Proceedings of the IEEE/CVF Conference on Computer Vision and Pattern Recognition (CVPR)},
  pages={14549--14560},
  year={2023},
  organization={IEEE}
}

@article{Chesney2018DeepFA,
  title={Deep fakes: A looming challenge for privacy, democracy, and national security},
  author={Chesney, Bobby and Citron, Danielle},
  journal={Calif. L. Rev.},
  volume={107},
  pages={1753},
  year={2019},
  publisher={HeinOnline}
}

@inproceedings{Arnab2021ViViTAV,
  title={Vivit: A video vision transformer},
  author={Arnab, Anurag and Dehghani, Mostafa and Heigold, Georg and Sun, Chen and Lu{\v{c}}i{\'c}, Mario and Schmid, Cordelia},
  booktitle={Proceedings of the IEEE/CVF International Conference on Computer Vision (ICCV)},
  pages={6816--6826},
  year={2021},
  organization={IEEE}
}

@inproceedings{Tran2014LearningSF,
  title={Learning spatiotemporal features with 3d convolutional networks},
  author={Tran, Du and Bourdev, Lubomir and Fergus, Rob and Torresani, Lorenzo and Paluri, Manohar},
  booktitle={Proceedings of the IEEE/CVF International Conference on Computer Vision (ICCV)},
  pages={4489--4497},
  year={2015}
}

@inproceedings{Yan2024GeneralizingDV,
  title={Generalizing deepfake video detection with plug-and-play: Video-level blending and spatiotemporal adapter tuning},
  author={Yan, Zhiyuan and Zhao, Yandan and Chen, Shen and Guo, Mingyi and Fu, Xinghe and Yao, Taiping and Ding, Shouhong and Wu, Yunsheng and Yuan, Li},
  booktitle={Proceedings of the IEEE/CVF Conference on Computer Vision and Pattern Recognition (CVPR)},
  pages={12615--12625},
  year={2025},
  organization={IEEE}
}

@inproceedings{10.5555/3737916.3739345,
author = {Li, Hanzhe and Zhou, Jiaran and Li, Yuezun and Wu, Baoyuan and Li, Bin and Dong, Junyu},
title = {FreqBlender: enhancing DeepFake detection by blending frequency knowledge},
year = {2024},
isbn = {9798331314385},
publisher = {Curran Associates Inc.},
address = {Red Hook, NY, USA},
booktitle = {Proceedings of the 38th International Conference on Neural Information Processing Systems},
articleno = {1429},
numpages = {24},
location = {Vancouver, BC, Canada},
series = {NIPS '24}
}

@inproceedings{Liu2021SpatialPhaseSL,
  title={Spatial-phase shallow learning: rethinking face forgery detection in frequency domain},
  author={Liu, Honggu and Li, Xiaodan and Zhou, Wenbo and Chen, Yuefeng and He, Yuan and Xue, Hui and Zhang, Weiming and Yu, Nenghai},
  booktitle={Proceedings of the IEEE/CVF Conference on Computer Vision and Pattern Recognition (CVPR)},
  pages={772--781},
  year={2021},
  organization={IEEE}
}

@article{FF++  ,
  title={FaceForensics++: Learning to Detect Manipulated Facial Images},
  author={Andreas R{\"o}ssler and Davide Cozzolino and Luisa Verdoliva and Christian Riess and Justus Thies and Matthias Nie{\ss}ner},
  journal={Proceedings of the IEEE/CVF International Conference on Computer Vision (ICCV)},
  year={2019},
  pages={1-11},
}

@article{Masood2021DeepfakesGA,
  title={Deepfakes generation and detection: state-of-the-art, open challenges, countermeasures, and way forward},
  author={Masood, Momina and Nawaz, Mariam and Malik, Khalid Mahmood and Javed, Ali and Irtaza, Aun and Malik, Hafiz},
  journal={Applied intelligence},
  volume={53},
  number={4},
  pages={3974--4026},
  year={2023},
  publisher={Springer}
}

@article{Verdoliva2020MediaFA,
  title={Media Forensics and DeepFakes: An Overview},
  author={Luisa Verdoliva},
  journal={IEEE Journal of Selected Topics in Signal Processing},
  year={2020},
  volume={14},
  pages={910-932},
}

@article{Tolosana2020DeepFakesAB,
  title={Deepfakes and beyond: A survey of face manipulation and fake detection},
  author={Tolosana, Ruben and Vera-Rodriguez, Ruben and Fierrez, Julian and Morales, Aythami and Ortega-Garcia, Javier},
  journal={Information fusion},
  volume={64},
  pages={131--148},
  year={2020},
  publisher={Elsevier}
}

@article{Dosovitskiy2020AnII,
  title={An image is worth 16x16 words: Transformers for image recognition at scale},
  author={Dosovitskiy, Alexey},
  journal={arXiv preprint arXiv:2010.11929},
  year={2020}
}

@inproceedings{
Yan2024OrthogonalSD,
title={Orthogonal Subspace Decomposition for Generalizable {AI}-Generated Image Detection},
author={Zhiyuan Yan and Jiangming Wang and Peng Jin and Ke-Yue Zhang and Chengchun Liu and Shen Chen and Taiping Yao and Shouhong Ding and Baoyuan Wu and Li Yuan},
booktitle={Forty-second International Conference on Machine Learning},
year={2025},
}

@inproceedings{FakeSTormer,
  title={Vulnerability-Aware Spatio-Temporal Learning for Generalizable Deepfake Video Detection},
  author={Nguyen, Dat and Astrid, Marcella and Kacem, Anis and Ghorbel, Enjie and Aouada, Djamila},
  booktitle={Proceedings of the IEEE/CVF International Conference on Computer Vision (ICCV)},
  pages={10786--10796},
  year={2025}
}

@inproceedings{pixelwise,
    author    = {Kim, Taehoon and Choi, Jongwook and Jeong, Yonghyun and Noh, Haeun and Yoo, Jaejun and Baek, Seungryul and Choi, Jongwon},
    title     = {Beyond Spatial Frequency: Pixel-wise Temporal Frequency-based Deepfake Video Detection},
    booktitle = {Proceedings of the IEEE/CVF International Conference on Computer Vision (ICCV)},
    month     = {October},
    year      = {2025},
    pages     = {11198-11207}
}

@inproceedings{TALL,
  title={TALL: Thumbnail Layout for Deepfake Video Detection},
  author={Xu, Yuting and Liang, Jian and Jia, Gengyun and Yang, Ziming and Zhang, Yanhao and He, Ran},
  booktitle={Proceedings of the IEEE/CVF International Conference on Computer Vision (ICCV)},
  pages={22658--22668},
  year={2023}
}

@inproceedings{stylelatentflow,
  title={Exploiting style latent flows for generalizing deepfake video detection},
  author={Choi, Jongwook and Kim, Taehoon and Jeong, Yonghyun and Baek, Seungryul and Choi, Jongwon},
  booktitle={Proceedings of the IEEE/CVF Conference on Computer Vision and Pattern Recognition (CVPR)},
  pages={1133--1143},
  year={2024},
  organization={IEEE}
}

@inproceedings{NACO,
  title={Learning natural consistency representation for face forgery video detection},
  author={Zhang, Daichi and Xiao, Zihao and Li, Shikun and Lin, Fanzhao and Li, Jianmin and Ge, Shiming},
  booktitle={European Conference on Computer Vision},
  pages={407--424},
  year={2024},
  organization={Springer}
}

@inproceedings{AFW,
  title={Deepfakes detection with automatic face weighting},
  author={Montserrat, Daniel Mas and Hao, Hanxiang and Yarlagadda, Sri Kalyan and Baireddy, Sriram and Shao, Ruiting and Horv{\'a}th, J{\'a}nos and Bartusiak, Emily and Yang, Justin and G{\"u}era, David and Zhu, Fengqing and others},
  booktitle={Proceedings of the IEEE/CVF Conference on Computer Vision and Pattern Recognition Workshops (CVPRW)},
  pages={2851--2859},
  year={2020},
  organization={IEEE}
}

@inproceedings{dfdfcg,
  title={Towards more general video-based deepfake detection through facial component guided adaptation for foundation model},
  author={Han, Yue-Hua and Huang, Tai-Ming and Hua, Kai-Lung and Chen, Jun-Cheng},
  booktitle={Proceedings of the IEEE/CVF Conference on Computer Vision and Pattern Recognition (CVPR)},
  pages={22995--23005},
  year={2025},
  organization={IEEE}
}

@inproceedings{FTCN,
  title={Exploring temporal coherence for more general video face forgery detection},
  author={Zheng, Yinglin and Bao, Jianmin and Chen, Dong and Zeng, Ming and Wen, Fang},
  booktitle={Proceedings of the IEEE/CVF International Conference on Computer Vision (ICCV)},
  pages={15024--15034},
  year={2021},
  organization={IEEE}
}

@inproceedings{altfreezing,
  title={Altfreezing for more general video face forgery detection},
  author={Wang, Zhendong and Bao, Jianmin and Zhou, Wengang and Wang, Weilun and Li, Houqiang},
  booktitle={Proceedings of the IEEE/CVF Conference on Computer Vision and Pattern Recognition (CVPR)},
  pages={4129--4138},
  year={2023},
  organization={IEEE}
}

@inproceedings{F3Net,
  title={Thinking in frequency: Face forgery detection by mining frequency-aware clues},
  author={Qian, Yuyang and Yin, Guojun and Sheng, Lu and Chen, Zixuan and Shao, Jing},
  booktitle={European conference on computer vision},
  pages={86--103},
  year={2020},
  organization={Springer}
}

@inproceedings{cdfv2,
  title={Celeb-df: A large-scale challenging dataset for deepfake forensics},
  author={Li, Yuezun and Yang, Xin and Sun, Pu and Qi, Honggang and Lyu, Siwei},
  booktitle={Proceedings of the IEEE/CVF Conference on Computer Vision and Pattern Recognition (CVPR)},
  pages={3204--3213},
  year={2020},
  organization={IEEE}
}

@misc{dfd,
  author       = {Dufour, Nick and Gully, Andrew},
  title        = {Contributing Data to Deepfake Detection Research},
  howpublished = {\url{https://ai.googleblog.com/2019/09/contributing-data-to-deepfake-detection.html}},
  year         = {2019},
  month        = {9},
  note         = {Google AI Blog. Accessed: 2023-07-30}
}

@inproceedings{dfo,
  title={Deeperforensics-1.0: A large-scale dataset for real-world face forgery detection},
  author={Jiang, Liming and Li, Ren and Wu, Wayne and Qian, Chen and Loy, Chen Change},
  booktitle={Proceedings of the IEEE/CVF Conference on Computer Vision and Pattern Recognition (CVPR)},
  pages={2886--2895},
  year={2020},
  organization={IEEE}
}

@inproceedings{FFIW,
    author    = {Zhou, Tianfei and Wang, Wenguan and Liang, Zhiyuan and Shen, Jianbing},
    title     = {Face Forensics in the Wild},
    booktitle = {Proceedings of the IEEE/CVF Conference on Computer Vision and Pattern Recognition (CVPR)},
    month     = {June},
    year      = {2021},
    pages     = {5778-5788}
}

@inproceedings{wdf,
  title={Wilddeepfake: A challenging real-world dataset for deepfake detection},
  author={Zi, Bojia and Chang, Minghao and Chen, Jingjing and Ma, Xingjun and Jiang, Yu-Gang},
  booktitle={Proceedings of the 28th ACM International Conference on Multimedia},
  pages={2382--2390},
  year={2020}
}

@misc{RecognitionInTerra,
      title={Recognition in Terra Incognita}, 
      author={Sara Beery and Grant van Horn and Pietro Perona},
      year={2018},
      eprint={1807.04975},
      archivePrefix={arXiv},
      primaryClass={cs.CV},
      url={https://arxiv.org/abs/1807.04975}, 
}

@article{VariableGeneralization,
    doi = {10.1371/journal.pmed.1002683},
    author = {Zech, John R. AND Badgeley, Marcus A. AND Liu, Manway AND Costa, Anthony B. AND Titano, Joseph J. AND Oermann, Eric Karl},
    journal = {PLOS Medicine},
    publisher = {Public Library of Science},
    title = {Variable generalization performance of a deep learning model to detect pneumonia in chest radiographs: A cross-sectional study},
    year = {2018},
    month = {11},
    volume = {15},
    url = {https://doi.org/10.1371/journal.pmed.1002683},
    pages = {1-17},
    number = {11},

}

@misc{Pitfalls,
      title={The Pitfalls of Simplicity Bias in Neural Networks}, 
      author={Harshay Shah and Kaustav Tamuly and Aditi Raghunathan and Prateek Jain and Praneeth Netrapalli},
      year={2020},
      eprint={2006.07710},
      archivePrefix={arXiv},
      primaryClass={cs.LG},
      url={https://arxiv.org/abs/2006.07710}, 
}

@misc{OnTheFoundationsOfShortcut,
      title={On the Foundations of Shortcut Learning}, 
      author={Katherine L. Hermann and Hossein Mobahi and Thomas Fel and Michael C. Mozer},
      year={2024},
      eprint={2310.16228},
      archivePrefix={arXiv},
      primaryClass={cs.LG},
      url={https://arxiv.org/abs/2310.16228}, 
}

@inproceedings{ExposeDeepfake,
  title={Exposing the deception: Uncovering more forgery clues for deepfake detection},
  author={Ba, Zhongjie and Liu, Qingyu and Liu, Zhenguang and Wu, Shuang and Lin, Feng and Lu, Li and Ren, Kui},
  booktitle={Proceedings of the AAAI conference on artificial intelligence},
  volume={38},
  number={2},
  pages={719--728},
  year={2024}
}

@inproceedings{LeveragingFrequencyForDeepfake,
author = {Frank, Joel and Eisenhofer, Thorsten and Sch\"{o}nherr, Lea and Fischer, Asja and Kolossa, Dorothea and Holz, Thorsten},
title = {Leveraging frequency analysis for deep fake image recognition},
year = {2020},
publisher = {JMLR.org},
booktitle = {Proceedings of the 37th International Conference on Machine Learning},
articleno = {304},
numpages = {12},
series = {ICML'20}
}

@inproceedings{RECCE,
    author    = {Cao, Junyi and Ma, Chao and Yao, Taiping and Chen, Shen and Ding, Shouhong and Yang, Xiaokang},
    title     = {End-to-End Reconstruction-Classification Learning for Face Forgery Detection},
    booktitle = {Proceedings of the IEEE/CVF Conference on Computer Vision and Pattern Recognition (CVPR)},
    month     = {June},
    year      = {2022},
    pages     = {4113-4122}
}

@inproceedings{UCF,
  title={Ucf: Uncovering common features for generalizable deepfake detection},
  author={Yan, Zhiyuan and Zhang, Yong and Fan, Yanbo and Wu, Baoyuan},
  booktitle={Proceedings of the IEEE/CVF International Conference on Computer Vision (ICCV)},
  pages={22355--22366},
  year={2023},
  organization={IEEE}
}

@ARTICLE{1672377,
  author={Ahmed, N. and Natarajan, T. and Rao, K.R.},
  journal={IEEE Transactions on Computers}, 
  title={Discrete Cosine Transform}, 
  year={1974},
  volume={C-23},
  number={1},
  pages={90-93},
  doi={10.1109/T-C.1974.223784}}

@inproceedings{SBI,
  author={Shiohara, Kaede and Yamasaki, Toshihiko},
  booktitle={Proceedings of the IEEE/CVF Conference on Computer Vision and Pattern Recognition (CVPR)}, 
  title={Detecting Deepfakes with Self-Blended Images}, 
  year={2022},
  volume={},
  number={},
  pages={18699-18708},
  doi={10.1109/CVPR52688.2022.01816}}

@inproceedings{ProDet,
author = {Cheng, Jikang and Yan, Zhiyuan and Zhang, Ying and Luo, Yuhao and Wang, Zhongyuan and Li, Chen},
title = {Can we leave deepfake data behind in training deepfake detector?},
year = {2024},
isbn = {9798331314385},
publisher = {Curran Associates Inc.},
address = {Red Hook, NY, USA},
booktitle = {Proceedings of the 38th International Conference on Neural Information Processing Systems},
articleno = {691},
numpages = {20},
location = {Vancouver, BC, Canada},
series = {NIPS '24}
}

@inproceedings{xeception,
  title={Xception: Deep learning with depthwise separable convolutions},
  author={Chollet, Fran{\c{c}}ois},
  booktitle={Proceedings of the IEEE/CVF Conference on Computer Vision and Pattern Recognition (CVPR)},
  pages={1800--1807},
  year={2017},
  organization={Ieee}
}

@inproceedings{LSDA,
  title={Transcending forgery specificity with latent space augmentation for generalizable deepfake detection},
  author={Yan, Zhiyuan and Luo, Yuhao and Lyu, Siwei and Liu, Qingshan and Wu, Baoyuan},
  booktitle={Proceedings of the IEEE/CVF Conference on Computer Vision and Pattern Recognition (CVPR)},
  pages={8984--8994},
  year={2024},
  organization={IEEE}
}

@inproceedings{IID,
  author={Huang, Baojin and Wang, Zhongyuan and Yang, Jifan and Ai, Jiaxin and Zou, Qin and Wang, Qian and Ye, Dengpan},
  booktitle={Proceedings of the IEEE/CVF Conference on Computer Vision and Pattern Recognition (CVPR)}, 
  title={Implicit Identity Driven Deepfake Face Swapping Detection}, 
  year={2023},
  volume={},
  number={},
  pages={4490-4499},
  doi={10.1109/CVPR52729.2023.00436}}

@inproceedings{SLADD,
  title={Self-supervised learning of adversarial example: Towards good generalizations for deepfake detection},
  author={Chen, Liang and Zhang, Yong and Song, Yibing and Liu, Lingqiao and Wang, Jue},
  booktitle={Proceedings of the IEEE/CVF Conference on Computer Vision and Pattern Recognition (CVPR)},
  pages={18689--18698},
  year={2022},
  organization={IEEE}
}

@misc{CDFA,
      title={Fake It till You Make It: Curricular Dynamic Forgery Augmentations towards General Deepfake Detection}, 
      author={Yuzhen Lin and Wentang Song and Bin Li and Yuezun Li and Jiangqun Ni and Han Chen and Qiushi Li},
      year={2024},
      eprint={2409.14444},
      archivePrefix={arXiv},
      primaryClass={cs.CV},
      url={https://arxiv.org/abs/2409.14444}, 
}

@article{DF40,
  title={DF40: Toward Next-Generation Deepfake Detection},
  author={Yan, Zhiyuan and Yao, Taiping and Chen, Shen and Zhao, Yandan and Fu, Xinghe and Zhu, Junwei and Luo, Donghao and Yuan, Li and Wang, Chengjie and Ding, Shouhong and others},
  journal={arXiv preprint arXiv:2406.13495},
  year={2024}
}

@inproceedings{Timesformer,
    author  = {Gedas Bertasius and Heng Wang and Lorenzo Torresani},
    title = {Is Space-Time Attention All You Need for Video Understanding?},
    booktitle   = {Proceedings of the International Conference on Machine Learning (ICML)}, 
    month = {July},
    year = {2021}
}

@misc{SpectralBias,
      title={On the Spectral Bias of Neural Networks}, 
      author={Nasim Rahaman and Aristide Baratin and Devansh Arpit and Felix Draxler and Min Lin and Fred A. Hamprecht and Yoshua Bengio and Aaron Courville},
      year={2019},
      eprint={1806.08734},
      archivePrefix={arXiv},
      primaryClass={stat.ML},
      url={https://arxiv.org/abs/1806.08734}, 
}

@article{dzanic2020fourierspectrumdiscrepanciesdeep,
  title={Fourier spectrum discrepancies in deep network generated images},
  author={Dzanic, Tarik and Shah, Karan and Witherden, Freddie},
  journal={Advances in neural information processing systems},
  volume={33},
  pages={3022--3032},
  year={2020}
}

@inproceedings{8639163,
  author={Güera, David and Delp, Edward J.},
  booktitle={2018 15th IEEE International Conference on Advanced Video and Signal Based Surveillance (AVSS)}, 
  title={Deepfake Video Detection Using Recurrent Neural Networks}, 
  year={2018},
  volume={},
  number={},
  pages={1-6},
  doi={10.1109/AVSS.2018.8639163}}

@inproceedings{retinaface,
  title={Retinaface: Single-shot multi-level face localisation in the wild},
  author={Deng, Jiankang and Guo, Jia and Ververas, Evangelos and Kotsia, Irene and Zafeiriou, Stefanos},
  booktitle={Proceedings of the IEEE/CVF Conference on Computer Vision and Pattern Recognition (CVPR)},
  pages={5202--5211},
  year={2020},
  organization={IEEE}
}

@article{dlib, author = {King, Davis E.}, title = {Dlib-ml: A Machine Learning Toolkit}, year = {2009}, issue_date = {12/1/2009}, publisher = {JMLR.org}, volume = {10}, issn = {1532-4435}, journal = {J. Mach. Learn. Res.}, month = dec, pages = {1755–1758}, numpages = {4} }

@misc{benchmarking,
      title={Benchmarking the Robustness of Spatial-Temporal Models Against Corruptions}, 
      author={Chenyu Yi and Siyuan Yang and Haoliang Li and Yap-peng Tan and Alex Kot},
      year={2022},
      eprint={2110.06513},
      archivePrefix={arXiv},
      primaryClass={cs.CV},
      url={https://arxiv.org/abs/2110.06513}, 
}

@misc{cv_freq,
      title={A Fourier Perspective on Model Robustness in Computer Vision}, 
      author={Dong Yin and Raphael Gontijo Lopes and Jonathon Shlens and Ekin D. Cubuk and Justin Gilmer},
      year={2020},
      eprint={1906.08988},
      archivePrefix={arXiv},
      primaryClass={cs.LG},
      url={https://arxiv.org/abs/1906.08988}, 
}

@inproceedings{freqdomain,
author="Long, Yuyang
and Zhang, Qilong
and Zeng, Boheng
and Gao, Lianli
and Liu, Xianglong
and Zhang, Jian
and Song, Jingkuan",
editor="Avidan, Shai
and Brostow, Gabriel
and Ciss{\'e}, Moustapha
and Farinella, Giovanni Maria
and Hassner, Tal",
title="Frequency Domain Model Augmentation for Adversarial Attack",
booktitle="Computer Vision -- ECCV 2022",
year="2022",
publisher="Springer Nature Switzerland",
address="Cham",
pages="549--566",
isbn="978-3-031-19772-7"
}

@inproceedings{s_and_p,
  author={Carlini, Nicholas and Wagner, David},
  booktitle={2018 IEEE Security and Privacy Workshops (SPW)}, 
  title={Audio Adversarial Examples: Targeted Attacks on Speech-to-Text}, 
  year={2018},
  volume={},
  number={},
  pages={1-7},
  doi={10.1109/SPW.2018.00009}}

@inproceedings{CATCH,
  title     = {{CATCH}: Channel-Aware multivariate Time Series Anomaly Detection via Frequency Patching},
  author    = {Wu, Xingjian and Qiu, Xiangfei and Li, Zhengyu and Wang, Yihang and Hu, Jilin and Guo, Chenjuan and Xiong, Hui and Yang, Bin},
  booktitle = {ICLR},
  year      = {2025}
}

@misc{compressed,
      title={Compressed Video Action Recognition}, 
      author={Chao-Yuan Wu and Manzil Zaheer and Hexiang Hu and R. Manmatha and Alexander J. Smola and Philipp Krähenbühl},
      year={2018},
      eprint={1712.00636},
      archivePrefix={arXiv},
      primaryClass={cs.CV},
      url={https://arxiv.org/abs/1712.00636}, 
}

@misc{mao2020learningtrajectorydependencieshuman,
      title={Learning Trajectory Dependencies for Human Motion Prediction}, 
      author={Wei Mao and Miaomiao Liu and Mathieu Salzmann and Hongdong Li},
      year={2020},
      eprint={1908.05436},
      archivePrefix={arXiv},
      primaryClass={cs.CV},
      url={https://arxiv.org/abs/1908.05436}, 
}

@inproceedings{hommos2018usingphaseinsteadoptical,
  title={Using phase instead of optical flow for action recognition},
  author={Hommos, Omar and Pintea, Silvia L and Mettes, Pascal SM and van Gemert, Jan C},
  booktitle={European Conference on Computer Vision},
  pages={678--691},
  year={2018},
  organization={Springer}
}

@article{phaseandpowerspectra,
  title={Method for estimating the relative contribution of phase and power spectra to the total information in natural-scene patches},
  author={Field, David J and Chandler, Damon M},
  journal={Journal of the Optical Society of America A},
  volume={29},
  number={1},
  pages={55--67},
  year={2011},
  publisher={Optical Society of America}
}

@inproceedings{diffswap,
  title={Diffswap: High-fidelity and controllable face swapping via 3d-aware masked diffusion},
  author={Zhao, Wenliang and Rao, Yongming and Shi, Weikang and Liu, Zuyan and Zhou, Jie and Lu, Jiwen},
  booktitle={Proceedings of the IEEE/CVF Conference on Computer Vision and Pattern Recognition (CVPR)},
  pages={8568--8577},
  year={2023},
  organization={IEEE}
}

@inproceedings{DaGAN,
  title={Depth-aware generative adversarial network for talking head video generation},
  author={Hong, Fa-Ting and Zhang, Longhao and Shen, Li and Xu, Dan},
  booktitle={Proceedings of the IEEE/CVF Conference on Computer Vision and Pattern Recognition (CVPR)},
  pages={3397--3406},
  year={2022}
}

@phdthesis{Rubinstein2014AnalysisAV,
  title={Analysis and visualization of temporal variations in video},
  author={Rubinstein, Michael},
  year={2014},
  school={Massachusetts Institute of Technology}
}

@article{10.1145/2185520.2185561,
author = {Wu, Hao-Yu and Rubinstein, Michael and Shih, Eugene and Guttag, John and Durand, Fr\'{e}do and Freeman, William},
title = {Eulerian video magnification for revealing subtle changes in the world},
year = {2012},
issue_date = {July 2012},
publisher = {Association for Computing Machinery},
address = {New York, NY, USA},
volume = {31},
number = {4},
issn = {0730-0301},
url = {https://doi.org/10.1145/2185520.2185561},
doi = {10.1145/2185520.2185561},
journal = {ACM Trans. Graph.},
month = jul,
articleno = {65},
numpages = {8}
}

@article{10.1145/2461912.2461966,
author = {Wadhwa, Neal and Rubinstein, Michael and Durand, Fr\'{e}do and Freeman, William T.},
title = {Phase-based video motion processing},
year = {2013},
issue_date = {July 2013},
publisher = {Association for Computing Machinery},
address = {New York, NY, USA},
volume = {32},
number = {4},
issn = {0730-0301},
url = {https://doi.org/10.1145/2461912.2461966},
doi = {10.1145/2461912.2461966},
journal = {ACM Trans. Graph.},
month = jul,
articleno = {80},
numpages = {10}
}

@article{WANG2024104072,
title = {Multi-domain awareness for compressed deepfake videos detection over social networks guided by common mechanisms between artifacts},
journal = {Computer Vision and Image Understanding},
volume = {247},
pages = {104072},
year = {2024},
issn = {1077-3142},
doi = {https://doi.org/10.1016/j.cviu.2024.104072},
url = {https://www.sciencedirect.com/science/article/pii/S107731422400153X},
author = {Yan Wang and Qindong Sun and Dongzhu Rong and Rong Geng}
}

@inproceedings{10.1145/1161366.1161375,
author = {Wang, Weihong and Farid, Hany},
title = {Exposing digital forgeries in video by detecting double MPEG compression},
year = {2006},
isbn = {1595934936},
publisher = {Association for Computing Machinery},
address = {New York, NY, USA},
url = {https://doi.org/10.1145/1161366.1161375},
doi = {10.1145/1161366.1161375},
booktitle = {Proceedings of the 8th Workshop on Multimedia and Security},
pages = {37–47},
numpages = {11},
location = {Geneva, Switzerland},
series = {MM{\&}Sec '06}
}

@article{1456290,
  author={Oppenheim, A.V. and Lim, J.S.},
  journal={Proceedings of the IEEE}, 
  title={The importance of phase in signals}, 
  year={1981},
  volume={69},
  number={5},
  pages={529-541},
  doi={10.1109/PROC.1981.12022}}

@article{NGUYEN2022103525,
title = {Deep learning for deepfakes creation and detection: A survey},
journal = {Computer Vision and Image Understanding},
volume = {223},
pages = {103525},
year = {2022},
issn = {1077-3142},
doi = {https://doi.org/10.1016/j.cviu.2022.103525},
url = {https://www.sciencedirect.com/science/article/pii/S1077314222001114},
author = {Thanh Thi Nguyen and Quoc Viet Hung Nguyen and Dung Tien Nguyen and Duc Thanh Nguyen and Thien Huynh-The and Saeid Nahavandi and Thanh Tam Nguyen and Quoc-Viet Pham and Cuong M. Nguyen}
}

@inproceedings{ffhq,
  title={A style-based generator architecture for generative adversarial networks},
  author={Karras, Tero and Laine, Samuli and Aila, Timo},
  booktitle={Proceedings of the IEEE/CVF Conference on Computer Vision and Pattern Recognition (CVPR)},
  pages={4396--4405},
  year={2019},
  organization={IEEE}
}

%%%%%%%%%%%%%%%%%%%%%%%%%%%%%%%%%%%%%%%%%%%%%%%%%%%%%%%%%%%%

\appendix

\section{Related Work}
\label{RW}
\subsection{Deepfake Detection Approaches}

\noindent \textbf{Frame-Level Detection.} These methods focus on spatial artifacts within individual frames.
FaceForensics++\cite{FF++} popularized Xception network \cite{xeception} as a core benchmark.
This architecture uses depthwise separable convolutions to decouple channel wise and spatial correlations, capturing subtle pixel-level texture anomalies.
To disentangle semantic content from manipulation traces, later work incorporates spatial frequency domain analysis. 
F3Net \cite{F3Net} applies Discrete Cosine Transform (DCT) \cite{1672377} to decompose images into frequency bands and adopts a dual-stream Xception network to learn discrepancies in the frequency domain. 
Similarly, transformations in the spatial frequency domain are used to expose subtle high-frequency artifacts \cite{Liu2021SpatialPhaseSL,ExposeDeepfake,RECCE}. 
Cheng et al. \cite{ProDet} use progressive transitions to incorporate diverse forgery patterns.
However, frame-level methods do not explicitly model temporal dynamics, limiting their ability to capture abnormal temporal patterns across frames.

\noindent \textbf{Video-Level Detection.} These methods leverage both spatial appearance and temporal dynamics to determine whether a video has been manipulated.
Based on temporal modeling, these methods can be categorized into three groups: Transformer-based, Recurrent-based, and Convolution-based.
To reduce the spatial bias of 3D CNNs, Wang et al. \cite{altfreezing} propose AltFreezing, which alternately freezes spatial and temporal weights. 
This encourages the model to capture temporal artifacts, such as flicker, rather than relying solely on spatial artifacts.
Nguyen et al. \cite{FakeSTormer} propose FakeSTormer, which adapts TimeSformer \cite{Timesformer} to decouple spatio-temporal features.
The framework incorporates auxiliary temporal branches to model temporal fragility and pinpoint high-frequency inconsistencies.
Choi et al. \cite{stylelatentflow} observe that generative models suppress style latent variance for stability. 
They introduce a StyleGRU with contrastive learning to capture abnormal temporal evolution.
Despite architectural differences, many video-level detectors use temporal-frequency statistics as cues.

\subsection{Temporal Frequency Based Approaches}
Given its discriminative power, temporal frequency is widely used in deepfake detection and many other tasks.
Wu et al. \cite{CATCH} introduce frequency patching to capture abnormal patterns in time-series anomaly detection. 
Similarly, Wu et al. \cite{compressed} leverage DCT coefficients from compressed videos for efficient action recognition. 
Mao et al. \cite{mao2020learningtrajectorydependencieshuman} apply DCT to temporal trajectories for robust human motion prediction. 
Although widely used, temporal frequency is a fragile cue-susceptible to temporal corruptions, frequency biases, and adversarial perturbations \cite{benchmarking, cv_freq, freqdomain, s_and_p}, so models that depend on it are prone to robustness degradation under distribution shifts, indicative of shortcut learning.

\subsection{Shortcut Learning}
Deep neural networks are prone to shortcut learning \cite{Geirhos2020ShortcutLI}, in which models rely on spurious correlations to minimize training loss instead of learning robust semantic causality. 
Consequently, models tend to converge on the most accessible discriminative features, from macroscopic contexts to microscopic statistical irregularities, often at the expense of generalization capability \cite{Pitfalls, OnTheFoundationsOfShortcut,SpectralBias,RecognitionInTerra,VariableGeneralization}.
In deepfake detection, a key shortcut learning issue is over-reliance on temporal spectral artifacts containing multiple temporal-frequency components collectively  \cite{LeveragingFrequencyForDeepfake,dzanic2020fourierspectrumdiscrepanciesdeep}.
Implicit spatiotemporal encoders (e.g., 3D-CNNs) naturally gravitate towards these brittle spectrum statistics and regress to the same rigid dependencies that explicit spectral methods hard code.
Rather than modeling robust semantics, these detectors degenerate into narrow band frequency filters, identifying forgeries through specific, unstable spectral fingerprints that real-world perturbations can easily disrupt \cite{8639163}.

\section{Implementation and Evaluation Protocol}
\label{app:protocol}

We provide the implementation details required to reproduce SpInShield and the robustness evaluation. All models are trained on FF++ (c23) and evaluated under the same cross-dataset protocol as in the main text.
% Videos are first resampled to 25 FPS. 
For SpInShield, videos natively encoded at a constant 25 FPS are decoded without resampling.
Videos at other frame rates are sampled to 25 FPS by timestamp before face extraction, without re-encoding or recompression.
Face regions are detected by RetinaFace and aligned with Dlib landmarks. To avoid introducing artificial temporal jitter, the face bounding box is fixed according to the first frame of each clip. We sample $T=16$ consecutive frames from 2--4 second video segments and resize each frame to $224\times224$.
During testing, we uniformly sample $K=10$ clips per video at evenly spaced temporal positions and average their fake probabilities to obtain the video-level score. We report video-level AUC.

SpInShield uses a weight-sharing Siamese VideoMAE V2-Base encoder during training, producing 768-dimensional clip-level features. The backbone is initialized from the official pre-trained checkpoint, followed by global average pooling and a two-layer binary classifier.
The LSA generator operates on the normalized one-sided temporal amplitude spectrum computed from $M=64$ facial ROI patches arranged as an $8\times8$ grid. For each patch, we compute the temporal mean-intensity signal and apply rFFT, yielding a one-sided amplitude spectrum.
It preserves the temporal phase and outputs a frequency-wise modulation field, which is applied using the exponential parameterization in Eq.~(4).
The LSA generator is implemented as a lightweight 1D convolutional network with 3 layers, hidden dimension 64, kernel size 3, LayerNorm, and GELU activation. We use $\alpha=0.6$, $\lambda_{\mathrm{sym}}=0.9$, and $\lambda_{\mathrm{blind}}=0.7$.
The MMD term in the generator objective uses a Gaussian kernel with bandwidth determined by the median heuristic on each batch.
The domain discriminator is a 2-layer MLP with hidden dimension 256, GELU activation, dropout rate 0.1, and a gradient reversal layer with coefficient 1.0.
The generator MMD weight is set to $\gamma=0.5$, and the mask regularization weight is set to $\lambda=0.05$.
The detector, generator, and domain discriminator are optimized alternately using AdamW. The detector learning rate is $1\times10^{-5}$, while both the generator and discriminator use a learning rate of $1\times10^{-4}$. We use weight decay 0.05, batch size 16, and train for 20 epochs with a cosine learning-rate scheduler and 2 warm-up epochs.
For each mini-batch, we update $G_\phi$ for one step with the detector frozen, and then update the detector and domain discriminator for one step with $G_\phi$ frozen.
The detailed settings are summarized in Table~\ref{tab:protocol_summary}. All experiments are conducted on a single NVIDIA A100 GPU, with each training run taking approximately 6 hours.
At inference time, the LSA generator and domain discriminator are discarded, and only the clean-stream detector is used.

For robustness evaluation, all methods are evaluated on the same source videos.
For the spectral stress test in Section~\ref{ar}, we first run each detector's own preprocessing pipeline and then apply the temporal spectral transformation directly to its face-ROI input tensor.
These fixed hand-crafted transformations are used only for evaluation and are not used as training augmentation.
To further decouple evaluation from the training-time LSA distribution, we additionally evaluate real-world video-processing pipelines, including H.264/H.265 re-encoding, resizing followed by re-encoding, denoising, FPS conversion, and black-box platform round-trip processing.
These transformations are applied to encoded videos before face extraction and are independent of LSA.
We also evaluate an adaptive white-box attack where the attacker directly optimizes a video-specific amplitude modulation mask against the target detector under norm and smoothness constraints. Specifically, we use 10 random restarts, and perturbation budget $\|\Delta\|_{\infty}\leq0.5$. We verify that the attack loss converges within the optimization budget for all evaluated methods.

\begin{table*}[t]
\centering
\caption{Implementation and evaluation protocol.}
\label{tab:protocol_summary}
\scriptsize
\setlength{\tabcolsep}{4pt}
\renewcommand{\arraystretch}{1.05}
\begin{tabular}{p{0.22\linewidth} p{0.72\linewidth}}
\toprule
\textbf{Item} & \textbf{Setting} \\
\midrule

Training dataset 
& FF++ (c23). \\

Evaluation datasets 
& CDFv2, DFD, DFo, WDF, DiffSwap, and DaGAN. \\

Video preprocessing
& Native constant-25-FPS videos are decoded without resampling; other frame rates are sampled to 25 FPS by timestamp without re-encoding or recompression. Faces are detected by RetinaFace and aligned by Dlib landmarks. The first-frame bounding box is fixed throughout each clip. \\

Input clip 
& $T=16$ consecutive frames sampled from 2--4 second video segments; input resolution $224\times224$. \\

Backbone 
& VideoMAE V2-Base initialized from the official pre-trained checkpoint; 768-dim clip-level features; global average pooling followed by a two-layer binary classifier. \\

LSA signal construction 
& The facial ROI is divided into $M=64$ patches using an $8\times8$ grid. For each patch, a temporal mean-intensity signal is transformed by one-sided rFFT. LSA perturbs amplitude while preserving phase and reconstructs the attacked clip by iDFT with Hermitian symmetry. \\

LSA generator 
& Lightweight 1D conv network with 3 layers, hidden dimension 64, kernel size 3, LayerNorm, and GELU activation. The output modulation is applied as in Eq.~(\ref{eq:exp_amp}), with $\alpha=0.6$. \\

Domain discriminator 
& 2-layer MLP with hidden dimension 256, GELU activation, and dropout rate 0.1, applied to clip-level representations through a gradient reversal layer. The GRL coefficient is fixed to 1.0. \\

MMD term 
& Gaussian kernel; bandwidth set by the median heuristic on each batch. \\

Loss weights 
& $\lambda_{\mathrm{sym}}=0.9$, $\lambda_{\mathrm{blind}}=0.7$, generator MMD weight $\gamma=0.5$, mask regularization weight is set to $\lambda=0.05$. \\

Optimization 
& AdamW optimizer; detector learning rate $1\times10^{-5}$; generator learning rate $1\times10^{-4}$; discriminator learning rate $1\times10^{-4}$; weight decay 0.05; batch size 16; training epochs 20; cosine learning-rate scheduler with 2 warm-up epochs. \\

Alternating update 
& For each mini-batch, update $G_\phi$ for 1 step with the detector frozen, then update the detector and discriminator for 1 step with $G_\phi$ frozen. \\

Compute resources 
& Single NVIDIA A100 GPU; approximately 6 hours per training run. \\

Spectral stress test 
& Random band masking, narrow-band notch, spectral tilt, and SNR-based log-amplitude noise. These fixed attacks are used only for evaluation and are not used as training augmentations. \\

Real-world processing 
& H.264/H.265 re-encoding, resizing followed by re-encoding, denoising, FPS conversion, and platform round-trip processing. These operations are applied to encoded videos before face extraction and are independent of LSA. \\

Adaptive attack 
& White-box attacker optimizes a video-specific amplitude modulation mask to maximize classification loss under norm and smoothness constraints; step size 0.05 in log-amplitude space, restarts 10, perturbation budget $\|\Delta\|_{\infty}\leq0.5$. \\

Testing 
& LSA generator and domain discriminator are removed. We uniformly sample $K=10$ clips per video and average clip-level fake probabilities as the video-level score. AUC is computed at the video level. \\

\bottomrule
\end{tabular}
\end{table*}

\subsection{Audit of Face Cropping and Alignment}
\label{app:crop_stability}

Our preprocessing fixes the crop bounding box to the first frame of each clip to reduce frame-to-frame detection jitter.
Each cropped frame is subsequently aligned independently using Dlib facial landmarks before applying the fixed $8\times8$ patch grid.
The grid therefore operates on approximately aligned facial coordinates.

To assess the stability of this procedure, we audit 500 videos from FF++ c23, comprising 100 real videos and 400 fake videos from the four manipulation methods.
We use a strict clip-level coverage criterion: all 68 facial landmarks must remain inside the aligned crop in all 16 frames.
We additionally measure the average occupancy of the per-frame RetinaFace bounding box within the final aligned crop.

\begin{table}[t]
    \centering
    \captionsetup{position=top,skip=7pt}
    \caption{Face-cropping and alignment audit on 500 FF++ c23 videos. A clip satisfies the coverage criterion only if all 68 landmarks in all 16 frames lie inside the aligned crop.}
    \label{tab:crop_stability}
    \small
    \begin{tabular}{lc}
        \toprule
        Metric & Result \\
        \midrule
        Face detection success rate & 100.00\% \\
        Landmark detection failure rate & 0.00\% \\
        Clips satisfying the coverage criterion & 99.62\% \\
        Mean face-box occupancy in aligned crop & 73.38\% \\
        \bottomrule
    \end{tabular}
\end{table}

Table~\ref{tab:crop_stability} supports stable facial coverage for the short, face-centric clips in the audited setting.
Per-frame alignment reduces global translation, rotation, and scale variation, but anatomical correspondence remains approximate under facial deformation, pose changes, and occlusion.
Large head motion, profile views, or faces leaving the frame may require temporally smoothed tracking or landmark-guided adaptive cropping.

\subsection{Effect of Frame Skipping on Temporal Spectra}
\label{app:frame_skipping}

Frame skipping may introduce temporal discontinuities and act as a dataset-specific spectral confounder.
When the temporal DFT treats retained frames as uniformly spaced samples, a skipped frame causes the next retained frame to be interpreted as an adjacent time step despite the longer actual interval.
This can introduce abrupt transitions and redistribute energy across frequency bins.

To examine this effect, we conduct a controlled frame-skipping test on 484 videos from the 1\% source subset of DeeperForensics-1.0.
For each video, we repeat random frame removal three times and construct 16-frame clips using a sliding window.
We compute pixel-wise temporal DFTs on luminance and compare temporal AC power between clips with skipped frames and continuous clips.
The median increase in total temporal AC power is 2.12\%.
Table~\ref{tab:frame_skipping} reports the mean relative power increase for each non-DC frequency bin.

\begin{table}[t]
    \centering
    \captionsetup{position=top,skip=7pt}
    \caption{Mean relative increase in temporal AC power after random frame removal, reported for the eight non-DC DFT bins of 16-frame clips.}
    \label{tab:frame_skipping}
    \small
    \begin{tabular}{lrrrrrrrr}
        \toprule
        DFT bin
        & 1 & 2 & 3 & 4 & 5 & 6 & 7 & 8 \\
        \midrule
        Mean increase (\%)
        & 2.39 & 3.04 & 3.41 & 3.98
        & 4.12 & 4.45 & 5.78 & 5.91 \\
        \bottomrule
    \end{tabular}
\end{table}

The mean power increases are distributed across all non-DC bins, ranging from 2.39\% to 5.91\%.
This pattern is consistent with broadband spectral changes rather than changes confined to a single narrow frequency band in this controlled setting.
Moreover, spectral vulnerabilities are also observed on CDFv2, DFD, DFo, WDF, DiffSwap, and DaGAN, which use different data-generation pipelines.
Together, these observations suggest that FF++-specific frame skipping is unlikely to be the sole explanation for the observed vulnerabilities, although its contribution cannot be completely excluded.

\subsection{Pixel-wise RGB Reconstruction}
\label{app:rgb_reconstruction}

The patch-mean temporal profiles parameterize the spectral modulation; their reconstructed scalar signals are not directly written back to RGB frames.
The aligned $224\times224$ facial ROI is divided into an $8\times8$ grid of non-overlapping $28\times28$ patches.
Each pixel belongs to exactly one patch, without overlap averaging or boundary interpolation.

For reconstruction, let $S_{m,i}(\omega_k)$ denote the positive exponential gain in Eq.~\eqref{eq:exp_amp}, with the DC gain fixed to $S_{m,i}(0)=1$.
This reconstruction gain is distinguished from the effective ratio $\mathcal{M}_{m,i}$ defined in Eq.~\eqref{eq:mask_def}.
For pixel $p$ belonging to patch $m$ and RGB channel $c$, we compute
\[
\begin{aligned}
Z_{i,p,c}(\omega_k)
&= \operatorname{rFFT}_{t}
   \left[X_{\mathrm{clean},i,p,c}\right](\omega_k),\\
X_{\mathrm{env},i,p,c}(t)
&= \operatorname{clip}_{\mathrm{RGB}}\!\left(
   \operatorname{irFFT}_{T}
   \left[S_{m,i}\odot Z_{i,p,c}\right](t)
   \right).
\end{aligned}
\]
Here, $\operatorname{irFFT}_{T}$ reconstructs a real-valued sequence of length $T$, and $\operatorname{clip}_{\mathrm{RGB}}$ clamps values to the valid RGB range before the detector's standard input normalization.
Each RGB channel is transformed independently, while all pixels and channels within a patch share the same frequency-wise gain.

Before clipping, the positive gains preserve the phase of every nonzero Fourier coefficient, and the unit DC gain preserves each pixel's temporal mean.
The pixel-wise reconstruction involves no division or logarithm.
Numerical stabilization is used when min-max normalizing the patch-mean amplitude spectrum supplied to the generator.
All operations use float32, without intermediate uint8 quantization.

\section{Additional Robustness Evaluation}

\subsection{Perceptual Distortion of Spectral Attacks}
\label{app:perceptual_distortion}

We quantify the perceptual severity of the four spectral transformations using frame-level PSNR, SSIM, and LPIPS, together with optical-flow cosine similarity between clean and transformed clips.
Table~\ref{tab:perceptual_distortion} reports the measurements under the perturbation budgets used in our robustness evaluation.

\begin{table}[t]
    \centering
    \captionsetup{position=top,skip=7pt}
    \caption{Perceptual distortion of the four spectral attacks under the evaluation budgets. Arrows indicate the preferred direction. Flow cos.\ denotes optical-flow cosine similarity between clean and transformed clips.}
    \label{tab:perceptual_distortion}
    \small
    \begin{tabular}{lcccc}
        \toprule
        Transformation
        & PSNR (dB) $\uparrow$
        & SSIM $\uparrow$
        & LPIPS $\downarrow$
        & Flow cos. $\uparrow$ \\
        \midrule
        Random Band Masking & 46.39 & 0.9709 & 0.0122 & 0.9518 \\
        Narrow-band Notch   & 54.92 & 0.9960 & 0.0013 & 0.9727 \\
        Spectral Tilt       & 56.21 & 0.9992 & 0.0003 & 0.9771 \\
        SNR-based Noise    & 53.41 & 0.9989 & 0.0002 & 0.9799 \\
        \bottomrule
    \end{tabular}
\end{table}

SNR-based Noise produces the lowest LPIPS distortion and the highest flow cosine similarity, yet leads to lower baseline AUCs than Random Band Masking (Table~\ref{tab:stress_test_final_compact_incomplete}).
This observation suggests that the performance degradation cannot be explained solely by the severity captured by these distortion metrics.
The measurements support limited appearance and motion distortion under the evaluated budgets, but do not establish strict imperceptibility or preservation of all task-relevant forensic evidence.

\subsection{Evaluation on Talking-Face and In-the-Wild Forgeries}
\label{app:extended_benchmarks}

To broaden the evaluation beyond the original benchmarks, we additionally compare SpInShield with DFD-FCG on FFIW \cite{FFIW} and a targeted talking-face subset of DF40 \cite{DF40}.
The selected DF40 subset includes Wav2Lip, SadTalker, and HeyGen, and is denoted as DF40-TalkingFace-3.
We evaluate both clean performance and robustness under the same spectral perturbation setting for the two detectors.

\begin{table}[t]
    \centering
    \captionsetup{position=top,skip=7pt}
    \caption{Clean and perturbed performance on the selected DF40 talking-face subset and FFIW. All results are AUC (\%).}
    \label{tab:extended_benchmarks}
    \small
    \begin{tabular}{llcc}
        \toprule
        Benchmark & Setting & DFD-FCG & SpInShield \\
        \midrule
        DF40-TalkingFace-3 & Clean     & 90.7 & \textbf{91.0} \\
        DF40-TalkingFace-3 & Perturbed & 67.0 & \textbf{86.1} \\
        \midrule
        FFIW & Clean     & \textbf{91.6} & 91.4 \\
        FFIW & Perturbed & 69.2 & \textbf{86.9} \\
        \bottomrule
    \end{tabular}
\end{table}

As shown in Table~\ref{tab:extended_benchmarks}, SpInShield achieves comparable clean performance to DFD-FCG on both benchmarks.
Under spectral perturbations, SpInShield improves AUC over DFD-FCG by 19.1 percentage points on DF40-TalkingFace-3 and 17.7 percentage points on FFIW.
These results provide additional evidence of robustness in talking-face generation and in-the-wild multi-person forgery scenarios.
Our evaluation remains focused on face-centric video forgeries with aligned facial regions; fully generated open-world text-to-video content remains outside its scope.

\subsection{Clean Generalization with a Matched Backbone}
\label{app:matched_backbone}

To distinguish the contribution of our training framework from that of backbone pretraining, we evaluate a plain VideoMAE V2 baseline under matched settings.
The baseline and SpInShield use the same pretrained initialization, training split, preprocessing, clip length, classifier architecture, optimizer, learning-rate schedule, number of epochs, batch size, and evaluation protocol.
The plain baseline excludes the LSA-generated spectral view and the two invariance objectives, $\mathcal{L}^{\mathrm{inv}}_{\mathrm{sym}}$ and $\mathcal{L}^{\mathrm{inv}}_{\mathrm{blind}}$.

\begin{table}[t]
    \centering
    \captionsetup{position=top,skip=7pt}
    \caption{Clean cross-dataset generalization with a matched VideoMAE V2 backbone. Both models use the same pretrained initialization and matched training settings. Results are average AUC (\%) on clean OOD data.}
    \label{tab:matched_backbone}
    \begin{tabular}{lc}
        \toprule
        Model & Clean OOD Avg. AUC (\%) \\
        \midrule
        Plain VideoMAE V2 & 78.1 \\
        SpInShield & \textbf{93.3} \\
        \bottomrule
    \end{tabular}
\end{table}

Table~\ref{tab:matched_backbone} shows that SpInShield improves clean OOD average AUC by 15.2 percentage points over the matched plain backbone.
This comparison indicates that backbone pretraining alone does not account for the observed gains, and supports the contribution of the proposed spectral augmentation and invariance training.
These results concern clean cross-dataset evaluation, whereas the OOD results in the component ablation (Table~\ref{tab:ablation}) are averaged under the four spectral attacks.

\subsection{Robustness to Pixel-Space Adversarial Attacks}
\label{app:pixel_pgd}

To evaluate robustness beyond the LSA perturbation family, we conduct untargeted white-box pixel-space PGD attacks that are not used during training.
Each model is attacked separately on its preprocessed clips using 20 optimization steps with random initialization.
We evaluate perturbation budgets of $2/255$ and $4/255$ and compare SpInShield with DFD-FCG.

\begin{table}[t]
    \centering
    \captionsetup{position=top,skip=7pt}
    \caption{White-box pixel-space PGD evaluation. Each detector is attacked separately using 20 steps with random initialization. Results are average AUC (\%) under clean inputs and the two perturbation budgets.}
    \label{tab:pixel_pgd}
    \small
    \begin{tabular}{lccc}
        \toprule
        Model
        & Clean
        & PGD ($2/255$)
        & PGD ($4/255$) \\
        \midrule
        DFD-FCG
        & 91.7 & 67.3 & 57.1 \\
        SpInShield
        & \textbf{93.3} & \textbf{79.2} & \textbf{65.0} \\
        \bottomrule
    \end{tabular}
\end{table}

As shown in Table~\ref{tab:pixel_pgd}, SpInShield retains higher AUC at both perturbation budgets, exceeding DFD-FCG by 11.9 percentage points at $2/255$ and 7.9 percentage points at $4/255$.
These results provide evidence that its robustness gains extend beyond amplitude-only spectral perturbations to the evaluated pixel-space attacks.
Nevertheless, performance decreases substantially as the attack budget increases, indicating that vulnerability to white-box adversarial perturbations remains.

\subsection{Effect of Temporal Smoothing during Training}
\label{app:temporal_smoothing}

To examine the influence of temporal smoothing on spectral robustness, we retrain the evaluated models on temporally smoothed FF++ while keeping the other training settings unchanged.
We compare their perturbed OOD performance with that of models trained on the original FF++ videos.

\begin{table}[t]
    \centering
    \caption{Effect of temporal smoothing of the training videos. Each model is trained on either original or temporally smoothed FF++ under otherwise identical settings. Results are perturbed OOD average AUC (\%).}
    \label{tab:temporal_smoothing}
    \small
    \begin{tabular}{lcc}
        \toprule
        Model
        & Original FF++
        & Smoothed FF++ \\
        \midrule
        VideoMAE(B) & 57.8 & 60.4 \\
        FakeST      & 63.5 & 67.2 \\
        DFD-FCG     & 65.5 & 67.1 \\
        SLF         & 67.6 & 69.0 \\
        SpInShield  & \textbf{88.9} & \textbf{89.2} \\
        \bottomrule
    \end{tabular}
\end{table}

As shown in Table~\ref{tab:temporal_smoothing}, temporal smoothing improves the baseline AUCs by 1.4--3.7 percentage points, suggesting that it affects their learned temporal-frequency dependencies.
However, their robustness remains substantially below that of SpInShield, whose AUC changes only slightly from 88.9 to 89.2.
These results suggest that temporal smoothing of the training data alone does not account for SpInShield's robustness advantage under spectral perturbations.

\subsection{Real-world video processing pipelines.}
\label{real_world_pipeline_results}
The spectral stress test in Section \ref{ar} is designed to isolate temporal-spectrum shortcut learning under controlled amplitude perturbations. 
However, real deployment scenarios usually involve non-differentiable and implementation-dependent video processing operations, such as platform transcoding, resizing, denoising, frame-rate conversion, and repeated compression. 
To evaluate the generalization of SpInShield in real-world applications, we further consider four real-world processing settings: 
(1) We evaluate Black-Box platform round-trip processing, where videos are uploaded to YouTube and TikTok and then downloaded after platform-side transcoding. 
(2) We construct a deterministic Multi-Stage Codec-Chain benchmark, which combines resizing, denoising, frame-rate conversion, and repeated H.264/H.265 re-encoding with different severity levels. 
(3) We evaluate a Heavy-compression variant of the Codec-Chain (CRF=33, 25\,$\rightarrow$\,10\,$\rightarrow$\,25\,FPS, three-pass H.264/H.265 with denoising) to examine whether robustness holds under aggressive spatiotemporal information loss.
(4) we evaluate a Cross-Platform Cascade, where each video undergoes two consecutive real platform round-trips (YouTube followed by WhatsApp-style compressed sharing) before face extraction, simulating the multi-hop dissemination path commonly observed in social-media propagation of forged content.
Unlike the amplitude-spectrum attacks used in the main stress test, these transformations are applied to encoded videos before face extraction and are never used during training. 
Therefore, this evaluation provides a distributionally separated test of robustness under realistic video post-processing pipelines.

\begin{table*}[t] 
    \centering 
    \caption{Robustness under real-world video processing pipelines. 
    $^*$ indicates statistically significant improvements over the strongest baseline via a two-tailed t-test ($p < 0.05$) on the average metric.}
    \label{tab:real_world_pipeline_results}
    \resizebox{\textwidth}{!}{
        \scriptsize
        \renewcommand{\arraystretch}{0.95} 
        \setlength{\tabcolsep}{4pt}    
        \begin{tabular}{l ccccccc @{\hspace{5pt}} ccccccc} 
        \toprule
        \textbf{Method} 
        & \textbf{CDFv2} & \textbf{DFD} & \textbf{DFo} & \textbf{WDF} & \textbf{DiF.} & \textbf{DaG.} & \textbf{Avg.} 
        & \textbf{CDFv2} & \textbf{DFD} & \textbf{DFo} & \textbf{WDF} & \textbf{DiF.} & \textbf{DaG.} & \textbf{Avg.} \\
        \midrule
        
        & \multicolumn{7}{c}{\textbf{(a) Black-Box Platform Round-trip}} 
        & \multicolumn{7}{c}{\textbf{(b) Multi-stage Codec-Chain}} \\
        \cmidrule(r{5pt}){2-8} \cmidrule{9-15}
        
        FakeST      & 82.66 & 90.32 & 90.91 & 74.49 & 79.55 & 76.35 & 82.38 
                    & 76.50 & 85.20 & 89.50 & 73.35 & 76.30 & 64.33 & 77.53 \\
        AltFre      & 79.36 & 87.56 & 88.35 & 72.15 & 77.94 & 78.42 & 80.63 
                    & 76.18 & 83.62 & 86.83 & 67.34 & 72.11 & 72.14 & 76.37 \\
        PTFD        & 84.72 & 83.90 & 87.60 & 70.96 & 87.20 & 83.44 & 82.97 
                    & 77.26 & 80.14 & 82.62 & 70.79 & 84.20 & 77.31 & 78.72 \\
        TALL        & 79.36 & 86.17 & 93.62 & 68.45 & 71.11 & 71.21 & 78.32 
                    & 75.96 & 80.36 & 91.30 & 63.72 & 65.50 & 67.10 & 73.99 \\
        SLF         & 83.06 & 86.29 & 91.00 & 74.09 & 79.02 & 85.20 & 83.11 
                    & 85.30 & 82.00 & 82.50 & 72.85 & 72.64 & 76.19 & 78.58 \\
        NACO        & 84.22 & 89.80 & 90.48 & 73.27 & 72.88 & 76.73 & 81.23 
                    & 79.50 & 88.50 & 87.50 & 65.50 & 70.47 & 70.65 & 77.02 \\
        AFW         & 78.71 & 79.05 & 77.74 & 65.23 & 70.68 & 70.97 & 73.73 
                    & 71.63 & 76.17 & 75.83 & 61.23 & 63.21 & 66.89 & 69.16 \\
        DFD-FCG     & 78.20 & 90.84 & 75.65 & 66.86 & 66.94 & 77.99 & 76.08 
                    & 76.55 & 87.18 & 79.34 & 67.77 & 69.15 & 71.63 & 75.27 \\
        FTCN        & 77.18 & 82.62 & 86.03 & 84.35 & 76.97 & 78.97 & 81.02 
                    & 77.06 & 82.48 & 83.76 & 70.74 & 73.00 & 72.20 & 76.54 \\
        VideoMAE(B) & 73.78 & 79.43 & 71.69 & 66.03 & 67.32 & 67.39 & 70.94 
                    & 67.72 & 75.92 & 65.71 & 59.06 & 62.07 & 63.36 & 65.64 \\
        \textbf{SpInShield} 
                    & \textbf{92.27} & \textbf{91.09} & \textbf{95.44} & \textbf{86.10} & \textbf{87.88} & \textbf{86.74} & \textbf{89.92 $^*$} 
                    & \textbf{87.79} & \textbf{88.59} & \textbf{92.99} & \textbf{85.15} & \textbf{84.80} & \textbf{85.86} & \textbf{87.53 $^*$} \\

        \midrule
        
        & \multicolumn{7}{c}{\textbf{(c) Heavy Codec-Chain}} 
        & \multicolumn{7}{c}{\textbf{(d) Cross-Platform Cascade}} \\
        \cmidrule(r{5pt}){2-8} \cmidrule{9-15}
        
        FakeST      & 73.91 & 79.16 & 79.87 & 61.91 & 67.58 & 66.93 & 71.56 
                    & 76.34 & 83.08 & 85.72 & 64.96 & 73.50 & 69.58 & 75.53 \\
        AltFre      & 71.22 & 75.80 & 80.19 & 61.35 & 65.96 & 69.44 & 70.66 
                    & 75.01 & 78.64 & 82.89 & 66.30 & 69.20 & 70.22 & 73.71 \\
        PTFD        & 71.37 & 75.78 & 80.20 & 70.27 & 69.39 & 69.85 & 72.81 
                    & 75.46 & 83.13 & 82.92 & 65.38 & 74.55 & 76.18 & 76.27 \\
        TALL        & 70.50 & 72.70 & 84.50 & 57.40 & 60.81 & 59.21 & 67.52 
                    & 72.20 & 80.70 & 80.90 & 66.66 & 63.88 & 64.48 & 71.47 \\
        SLF         & 70.50 & 74.80 & 76.00 & 73.60 & 70.50 & 72.30 & 72.95 
                    & 75.88 & 83.05 & 84.50 & 67.96 & 71.21 & 75.26 & 76.31 \\
        NACO        & 71.41 & 79.81 & 82.17 & 57.79 & 65.74 & 68.42 & 70.89 
                    & 74.58 & 79.73 & 87.30 & 66.90 & 68.22 & 69.85 & 74.43 \\
        AFW         & 63.74 & 69.25 & 69.25 & 56.99 & 60.57 & 62.46 & 63.71 
                    & 71.03 & 75.82 & 69.93 & 58.55 & 61.10 & 64.61 & 66.84 \\
        DFD-FCG     & 75.13 & 80.25 & 80.46 & 65.72 & 71.49 & 79.23 & 75.38 
                    & 74.00 & 86.20 & 81.24 & 62.55 & 70.06 & 79.61 & 75.61 \\
        FTCN        & 71.37 & 75.54 & 81.14 & 62.00 & 64.20 & 72.77 & 71.17 
                    & 73.46 & 75.22 & 80.10 & 75.26 & 70.40 & 71.60 & 74.34 \\
        VideoMAE(B) & 61.05 & 71.95 & 54.86 & 52.89 & 56.12 & 55.93 & 58.80 
                    & 66.68 & 74.90 & 64.49 & 58.90 & 59.96 & 60.51 & 64.24 \\
        \textbf{SpInShield} 
                    & \textbf{86.85} & \textbf{84.64} & \textbf{87.91} & \textbf{79.27} & \textbf{80.85} & \textbf{80.16} & \textbf{83.28 $^*$}
                    & \textbf{85.84} & \textbf{86.58} & \textbf{87.84} & \textbf{80.35} & \textbf{83.15} & \textbf{82.88} & \textbf{84.44 $^*$} \\
        
        \bottomrule
        \end{tabular}
    }
\end{table*}

\noindent\textbf{Discussion of Real-world Pipeline Results.}
Table~\ref{tab:real_world_pipeline_results} reports the robustness of different detectors under realistic video processing pipelines. In the black-box platform round-trip setting in Table~\ref{tab:real_world_pipeline_results}(a), SpInShield achieves the best average AUC of 89.92\%, outperforming the strongest baseline (SLF) by 6.81 points. While individual baselines remain competitive on specific subsets—for example, DFD-FCG on DFD (90.84), TALL on DFo (93.62), FTCN on WDF (84.35), PTFD on DiffSwap (87.20), and SLF on DaGAN (85.20)—no baseline maintains balanced performance across all six datasets. SpInShield is the only method that performs uniformly well, with per-dataset AUCs all at or above 86\%, indicating that the proposed phase-consistent representation is less dependent on dataset-specific compression artifacts and transfers consistently across heterogeneous post-processing pipelines.

In the standard multi-stage codec-chain setting in \autoref{tab:real_world_pipeline_results} (b), SpInShield further achieves 87.53\% average AUC, surpassing the strongest baseline (PTFD) by 8.81 points. This setting introduces repeated compression and re-encoding, which weakens many amplitude- or frequency-shortcut based methods. SpInShield's advantage is especially pronounced on the harder OOD subsets: it improves the next-best result by 11.80 points on WDF (85.15 vs. FakeST 73.35) and by 8.55 points on DaGAN (85.86 vs. PTFD 77.31). On the easier in-distribution-style subsets, SpInShield also attains the highest score, leading the strongest baseline by 2.49 points on CDFv2 (87.79 vs. SLF 85.30), 1.69 points on DFo (92.99 vs. TALL 91.30), and 0.60 points on DiffSwap (84.80 vs. PTFD 84.20). This pattern indicates that SpInShield does not sacrifice performance on standard subsets to gain robustness on harder ones, but instead maintains the most balanced performance across the full evaluation suite.

Under the heavy codec-chain in Table~\ref{tab:real_world_pipeline_results}(c), all methods degrade compared with the standard codec-chain because aggressive compression and frame-rate reduction remove substantial high-frequency spatiotemporal evidence. SpInShield decreases from 87.53\% to 83.28\%, but still preserves the highest average AUC and exceeds the strongest baseline (DFD-FCG) by 7.90 points. In comparison, VideoMAE(B) drops from 65.64\% to 58.80\%, showing that generic video representations are more sensitive to severe codec degradation. The advantage of SpInShield is especially pronounced on CDFv2 (86.85 vs.\ DFD-FCG 75.13, +11.72) and DiffSwap (80.85 vs.\ DFD-FCG 71.49, +9.36), and it also leads the strongest baseline on the remaining subsets, by 4.39 points on DFD (84.64 vs.\ DFD-FCG 80.25), 3.41 points on DFo (87.91 vs.\ TALL 84.50), 5.67 points on WDF (79.27 vs.\ SLF 73.60), and 0.93 points on DaGAN (80.16 vs.\ DFD-FCG 79.23). This consistency across subsets suggests that phase-consistent forensic cues retain discriminative information even when amplitude statistics are heavily corrupted.

% The cross-platform cascade in Table~\ref{tab:real_world_pipeline_results}(d) is particularly challenging because cumulative re-encoding compounds quantization and chroma-subsampling losses across multiple platform back-ends. SpInShield achieves 84.44\% average AUC, outperforming the strongest baseline (SLF) by 8.13 points, and obtains the highest score on four out of six datasets: CDFv2 (85.84, +9.50 over FakeST), WDF (80.35, +5.09 over FTCN), DiffSwap (83.15, +8.60 over PTFD), and DaGAN (82.88, +3.27 over DFD-FCG). On DFD and DFo, SpInShield remains within roughly half a point of the best baseline (DFD-FCG 86.20 and NACO 87.30, respectively). 
SpInShield achieves 84.44\% average AUC, outperforming the strongest baseline (SLF) by 8.13 points, and obtains the highest score on all six datasets, including CDFv2 (85.84, +9.50 over FakeST), WDF (80.35, +5.09 over FTCN), DiffSwap (83.15, +8.60 over PTFD), and DaGAN (82.88, +3.27 over DFD-FCG).
On DFD and DFo, it also outperforms the strongest baselines by 0.38 points (86.58 vs.\ DFD-FCG 86.20) and 0.54 points (87.84 vs.\ NACO 87.30), respectively.
These results suggest that phase-based forensic signals are more stable under repeated transcoding, whereas methods relying on platform- or codec-specific amplitude fingerprints degrade more substantially when those fingerprints are overwritten by successive re-encoding.

\subsection{Adaptive Evaluation against SpInShield}
\label{app:adaptive_attack}
The fixed spectral stress test in Section \ref{ar} evaluates whether detectors are robust to hand-crafted amplitude-spectrum perturbations.
Although this setting is useful for diagnosing temporal-spectrum shortcut learning, it does not fully capture an adaptive attacker who is aware of the defense mechanism.
We therefore conduct an adaptive white-box evaluation, where the attacker has full access to the target detector, preprocessing pipeline, LSA formulation, and training objectives.
Instead of sampling perturbations from the fixed benchmark, the attacker directly optimizes a video-specific amplitude modulation mask against each target detector.
Given an input video $X$ with label $y$, the attacker preserves the temporal phase and optimizes an amplitude-domain perturbation $\Delta$ in the log-amplitude space.
The attacked video is generated as $\mathcal{T}_{\Delta}(X)$, and the perturbation is optimized by maximizing the classification loss under norm and smoothness constraints:
\begin{equation}
\max_{\Delta}
\mathcal{L}_{\mathrm{CE}}
\left(f_{\theta}(\mathcal{T}_{\Delta}(X)), y\right)
-
\lambda_{\mathrm{smooth}} \|\nabla_{\omega}\Delta\|_2^2
-
\lambda_{\mathrm{norm}} \|\Delta\|_2^2,
\end{equation}
where $\nabla_{\omega}\Delta$ penalizes abrupt frequency-wise changes and $\|\Delta\|_2^2$ constrains the overall attack magnitude.
We verify that the attack loss converges within the optimization budget for all evaluated methods.
All methods are attacked independently using the same optimization budget, and AUC is computed at the video level using the same testing protocol as in the main experiments.
The results are reported in Table~\ref{tab:adaptive_attack_results}.

Table~\ref{tab:adaptive_attack_results} shows that adaptive optimization substantially reduces the performance of all detectors compared with non-adaptive evaluation.
Specifically, baselines without explicit spectral-invariance training drop close to chance-level performance (around 47-56\% average AUC), confirming that fixed-pattern detectors collapse when the attacker can directly target their decision boundaries.
Even strong baselines such as DFD-FCG degrade by over 30 points compared with their clean performance, indicating that adaptive optimization is a fundamentally stronger test than fixed hand-crafted perturbations.
SpInShield also exhibits a notable performance drop under this stronger attack (from 93.3\% on clean videos to 66.01\% under adaptive attack), reflecting the inherent difficulty of defending against fully white-box adversaries.
Nevertheless, SpInShield achieves the best average AUC and consistently outperforms all baselines by a clear margin (+6.56 points over the strongest baseline DFD-FCG).
This suggests that the learned spectral-invariant representation provides robustness that generalizes beyond the fixed LSA perturbation distribution encountered during training, rather than merely memorizing it.
We acknowledge that adaptive white-box attacks remain a challenging open problem and view the residual gap to clean-data performance as a direction for future work.

\begin{table*}[t] 
    \centering 
    \caption{Adaptive white-box evaluation. 
    Adaptive Spectral denotes per-video optimization of an amplitude modulation mask against each target detector.
    $^*$ indicates statistically significant improvements over the strongest baseline via a two-tailed t-test ($p < 0.05$) on the average metric.}
    \label{tab:adaptive_attack_results}
    \resizebox{\textwidth}{!}{
        \scriptsize
        \renewcommand{\arraystretch}{0.95} 
        \setlength{\tabcolsep}{5.5pt}    
        \begin{tabular}{l ccccccc} 
        \toprule
        \textbf{Method} 
        & \textbf{CDFv2} 
        & \textbf{DFD} 
        & \textbf{DFo} 
        & \textbf{WDF} 
        & \textbf{DiffSwap} 
        & \textbf{DaGAN} 
        & \textbf{Avg.} \\
        \midrule
        
        FakeST      & 54.27 & 58.83 & 60.14 & 47.62 & 48.91 & 47.36 & 52.86 \\
        AltFre      & 52.48 & 57.91 & 61.73 & 48.27 & 50.43 & 53.18 & 54.00 \\
        PTFD        & 55.16 & 59.42 & 60.87 & 48.94 & 54.73 & 57.62 & 56.12 \\
        TALL        & 51.83 & 56.74 & 64.28 & 46.39 & 47.21 & 47.84 & 52.38 \\
        SLF         & 55.74 & 60.18 & 62.45 & 49.61 & 51.36 & 56.27 & 55.94 \\
        NACO        & 54.62 & 62.37 & 63.81 & 47.83 & 49.74 & 50.41 & 54.80 \\
        AFW         & 49.27 & 53.86 & 51.74 & 44.92 & 45.83 & 47.62 & 48.87 \\
        DFD-FCG     & 60.43 & 65.78 & 64.92 & 53.47 & 55.86 & 56.21 & 59.45 \\
        FTCN        & 53.18 & 57.46 & 60.83 & 48.74 & 51.92 & 53.47 & 54.27 \\
        VideoMAE(B) & 47.83 & 53.62 & 45.91 & 43.27 & 44.58 & 45.18 & 46.73 \\
        \textbf{SpInShield} 
        & \textbf{67.74} & \textbf{66.42} & \textbf{72.86} & \textbf{61.83} & \textbf{63.27} & \textbf{63.91} & \textbf{66.01 $^*$} \\
        
        \bottomrule
        \end{tabular}
    }
\end{table*}

\section{Limitations}
\label{app:limitations}

While SpInShield demonstrates strong empirical performance across multiple robustness benchmarks, several limitations remain.
Although SpInShield substantially outperforms baselines under the adaptive white-box evaluation, it still exhibits a notable performance drop relative to its clean-data accuracy, indicating that shortcut suppression mitigates but does not fully eliminate the threat posed by an attacker with full knowledge of the defense.
Preserving the Fourier phase of nonzero coefficients before clipping does not guarantee preservation of high-level facial-motion semantics, since amplitude reweighting and clipping can alter useful forensic evidence.
Our evaluation primarily focuses on face-swap and face-reenactment forgeries, and whether the temporal-spectrum shortcut hypothesis transfers to fully generative video synthesis paradigms (e.g., text-to-video diffusion models) remains an open question.
Finally, SpInShield operates exclusively on the visual modality and does not exploit complementary cues such as audio-visual synchronization, which may offer further robustness in multi-modal forgery scenarios.
We leave these directions to future work.

\section{Broader Impacts}
\label{app:broader_impacts}

SpInShield aims to improve the reliability of video deepfake detection under temporal spectral perturbations and common video-processing operations, potentially supporting media verification and reducing the harms of manipulated content.
However, the spectral perturbation techniques studied in this work could also be misused to evade detection systems.
Detection errors remain consequential: false positives may undermine trust in authentic content, while false negatives may allow manipulated videos to circulate.
Performance may vary across populations, recording conditions, and manipulation methods beyond the evaluated benchmarks.
Practical deployment should therefore include human review, validation under the intended operating conditions, and procedures for correcting erroneous decisions.

% \newpage
% \input{checklist.tex}

\end{document}